\documentclass[11pt]{article}
\usepackage[T1]{fontenc}
\usepackage[english]{babel}
\usepackage{CJKutf8} 
\usepackage{units}
\usepackage{amsthm}
\usepackage[flushleft]{threeparttable}
\usepackage{rotating}
\usepackage{arydshln}
\usepackage{amsmath}
\usepackage[space]{grffile}
\usepackage{graphicx}
\usepackage{setspace}
\usepackage{esint}
\usepackage{textcomp}
\usepackage{float}
\usepackage{apacite}
\usepackage{multirow}
\usepackage{rotfloat}
\usepackage{xr-hyper}
\usepackage{booktabs}
\usepackage{calc}
\usepackage{CJK}
\usepackage{psfrag,color} 
\usepackage{nicefrac} 
\usepackage{amsfonts} 
\usepackage{subfig} 
\usepackage{ae} 
\usepackage{aecompl} 
\usepackage{pifont} 
\usepackage{natbib} 
\usepackage{hyperref} 
\usepackage{soul}
\usepackage{lscape}
\usepackage{indentfirst}
\usepackage{rotating}
\usepackage[a4paper]{geometry}
\usepackage[flushleft]{threeparttable}
\makeatletter
\theoremstyle{plain}
  
\theoremstyle{plain}

  \theoremstyle{plain}
  
    \theoremstyle{plain}
  
\usepackage{listings}
\usepackage{listings}
\usepackage{xcolor}

\definecolor{codegreen}{rgb}{0,0.6,0}
\definecolor{codegray}{rgb}{0.5,0.5,0.5}
\definecolor{codepurple}{rgb}{0.58,0,0.82}
\definecolor{backcolour}{rgb}{0.95,0.95,0.92}

\lstdefinestyle{mystyle}{
    backgroundcolor=\color{backcolour},   
    commentstyle=\color{codegreen},
    keywordstyle=\color{magenta},
    numberstyle=\tiny\color{codegray},
    stringstyle=\color{codepurple},
    basicstyle=\ttfamily\footnotesize,
    breakatwhitespace=false,         
    breaklines=true,                 
    captionpos=b,                    
    keepspaces=true,                 
    numbers=left,                    
    numbersep=5pt,                  
    showspaces=false,                
    showstringspaces=false,
    showtabs=false,                  
    tabsize=2
}

\title{Selecting Between BERT and GPT for Text Classification in Political Science Research} 

\author{Yu Wang\footnote{yuwang.aiml@gmail.com}, Wen Qu, Xin Ye\\ Fudan University}

\date{}

\begin{document}

\maketitle

\begin{abstract}

\noindent Political scientists often grapple with data scarcity in text classification. Recently, fine-tuned BERT models and their variants have gained traction as effective solutions to address this issue. In this study, we investigate the potential of GPT-based models combined with prompt engineering as a viable alternative. We conduct a series of experiments across various classification tasks, differing in the number of classes and complexity, to evaluate the effectiveness of BERT-based versus GPT-based models in low-data scenarios. Our findings indicate that while zero-shot and few-shot learning with GPT models provide reasonable performance and are well-suited for early-stage research exploration, they generally fall short — or, at best, match — the performance of BERT fine-tuning, particularly as the training set reaches a substantial size (e.g., 1,000 samples). We conclude by comparing these approaches in terms of performance, ease of use, and cost, providing practical guidance for researchers facing data limitations. Our results are particularly relevant for those engaged in quantitative text analysis in low-resource settings or with limited labeled data.

\end{abstract}



\doublespacing
\setlength{\parindent}{2em}
\section{Introduction}
\noindent Text classification is one of the most common tasks in quantitative text analysis. Researchers often need to classify different texts into topics. Such texts encompass news articles~\citep{bert_nli,interpretable,automated_text_analysis,polarization_trump_clinton_followers,tweets_china}, tweets~\citep{german_compare_emotion,rhetoric_on_twitter}, public speeches~\citep{german_compare_emotion,pretrained_topic_classification}, video descriptions~\citep{youtube_ideology}, names~\citep{fuzzy,all_in_the_name}, among others. Regardless of the specific text form, one of the key bottlenecks in performing text classification is data scarcity: procuring labeled data is a slow and labor-intensive process and as a result the labeled set is oftentimes fairly small.

To resolve the data scarcity issue, researchers have explored various approaches. For example, some researchers have trained models using labeled cross-domain data, which is abundant, and then applied the trained model to in-domain classification~\citep{cross_domain}. Others have studied the plausibility of using ChatGPT as an automatic annotator to replace human annotation and speed up the labeling process~\citep{zero-shot}. Still others have considered instead of random sampling how to select more informative samples for labeling so as to reduce the number of labeled samples~\citep{sample_selection}. Thus far, however, the most effective approach has been finetuning BERT models~\citep{bert}. By coupling general knowledge in the pretrained language models and a few hundred task-specific samples, finetuned BERT models have proven to offer superior performance as compared with classical models such as logistic regression~\citep{bert_nli,finetune_pa}. Over the past few years, this pretrain-finetune paradigm~\citep{tutorial} has quickly established itself as the go-to method for text classification~\citep{bert_nli,youtube_ideology}.

In this article, we study zero-shot and few-shot prompting with GPT models as a potential alternative solution to the data scarcity problem.\footnote{Besides prompting, another alternative is to finetune these GPT models (https://platform.openai.com/docs/guides/fine-tuning/). We do not explore this approach here because our early explorations in this direction did not yield promising results.} Specifically, we analyze in situations with 1,000 or fewer samples how prompting with GPT models compares with finetuned BERT models in binary and multi-class classification. Through extensive experiments, we demonstrate that zero-shot and few-shot learning with GPT-based large language models can serve as an effective alternative to fine-tuning BERT models. The advantages of GPT models for classification are particularly prominent when the number of classes is small, e.g., 2, and when the task is easier.




\section{Text Classification in Political Science}
\noindent Quantitative text analysis has gone through quite a few distinctive methodological stages throughout its evolution: feature-based classical models, word embeddings, BERT models, and more recently generative models.\footnote{Some researchers have grouped the stages of word embeddings and BERT models into a unified representation learning stage~\citep{qta}.} As in other social science disciplines~\citep{qta,beyond_rating_scales,bag_of_words}, text analysis in political science research has followed a similar trajectory. Initially, researchers converted texts into counts and trained classical models from scratch. Subsequently, word counts were replaced with word embeddings, and recurrent neural networks were employed for classification. More recently, there has been a growing body of literature focused on fine-tuning BERT models.

\subsection{Classical Models}
\noindent Classical models refer mostly to the simpler and smaller models that take word counts as input. Naive Bayes, support vector machine and logistic regression models generally fall into this category~\citep{elements,bag_of_words,bestvater_monroe_2023}.\footnote{In addition to their use as natural language processing tools, these models are also often trained on tabular data. See for~\citep{predicted_acc1} and~\citep{predicted_acc2} for recent examples.} They are simpler in model architecture, smaller in model size, require training from scratch, and take word frequencies as input. Given that the order of words is not utilized, these models are considered as a ``bag of words'' approach. Other hallmarks of classical models include preprocessing and feature engineering. Because of the significance of word frequencies, careful preprocessing steps are usually required~\citep{embedding_discussion}. Given the number of words (i.e., features) can be enormous, researchers need to manually decide what features to include and what to exclude~\citep{bag_of_words}. Prominent examples that utilize classical models for text classification include~\cite{wheat_chaff}, which uses support vector machine to classify documents on the Militarized Interstate Dispute 4 (MID4) data collection project, and~\cite{language_ideology}, which uses support vector machine to classify U.S. senators into `(extreme) conservative' and `(extreme) liberal' using those senators' speeches as text input.


\subsection{Word Embeddings}
\noindent Word embeddings are vector representations of words~\citep{word2vec,glove,pca,embedding_discussion}. By projecting words into a vector space based on their co-occurrence patterns, word embeddings possess semantic meaning, with semantically similar words located close to each other in the vector space. Researchers have utilized word embeddings to study various topics, such as ideological placement in parliamentary corpora ~\citep{ideological_placement} and the evolving meaning of political concepts~\citep{political_concepts}. Beyond serving as standalone entities, these embeddings can also function as input to recurrent neural networks for text classification~\citep{lstm_pa,polarization_trump_clinton_followers}. For notable applications of embeddings in other social studies, readers can refer to~\cite{linguistic_agency},~\cite{promotional_language},~\cite{language_marker} and~\cite{depression_prediction}. 



\subsection{BERT Models}
\noindent BERT models, which are encoder-based language models, have emerged as one of the most effective tools for text classification~\citep{tutorial}. They are rooted in the transformer architecture first introduced by~\cite{attention}. Since their introduction in 2018~\citep{bert}, BERT models have consistently achieved state-of-the-art performance across various natural language processing tasks~\citep{bert}. Building on their initial success, numerous variations of BERT models have been developed, incorporating more extensive training data~\citep{roberta}, specialized data domains~\citep{conflibert,biobert}, and novel pretraining tasks~\citep{albert}. In the last couple of years, they have started to gain traction in political science research. Whether it is classifying news articles into different economic sentiments~\citep{bert_nli}, parliamentary speeches into different topics~\citep{pretrained_topic_classification}, tweets for depression detection~\citep{monitor_depression} or video descriptions into ideology categories~\citep{youtube_ideology}. In addition to their effectiveness as classifiers, BERT models have also been utilized by researchers for embedding tasks~\citep{promotional_language,language_marker,sample_selection,text_package}. Researchers first transform texts into embeddings using BERT models and then apply these embeddings in classical models such as logistic regression~\citep{embedding_discussion}.

\subsection{GPT Models}
\noindent GPT models are decoder-based language models and represent another highly effective tool for text analysis. Like BERT models, they also trace their roots back to the transformers introduced in~\cite{attention}. Unlike BERT models, GPT models are primarily designed for text generation. They excel in tasks such as essay writing, text summarization, translation, question answering, idea generation, and medical report transformation, among others~\citep{radford2019language,genai_for_economists,gpt4_radiology_reports_transformation}. Researchers in various fields, such as economics~\citep{turing_test_ai} and psychology~\citep{theory_of_mind}, have sought to leverage the generative capabilities of these models, exploring whether they behave similarly to humans in classical games like the ultimatum bargaining game and the prisoner's dilemma game. Similarly, political scientists have explored using GPT models to simulate human samples~\citep{argyle_busby_fulda_gubler_rytting_wingate_2023,synthetic_replacement}, investigating whether ``silicon samples'' respond to surveys in a manner akin to humans after the models have been conditioned on thousands of socio-demographic backstories from real participants. Others have explored leveraging these models' generative capabilities for chat interventions to improve online political conversations~\citep{democratic_discourse}.

In addition to the original text generation capabilities, as GPT models grow in size, they have started to demonstrate emergent abilities~\citep{emerging_capabilities} unseen in smaller model versions.\footnote{Per definition, ``an ability is emergent if it is not present in smaller models but is present in larger models.''~\citep{emerging_capabilities}} Among these emergent abilities are zero-shot prompting~\citep{radford2019language} and few-shot prompting~\citep{gpt3}. For example, researchers have explored the plausibility of using ChatGPT as an automatic annotator to replace human annotators~\citep{zero-shot}. Others have studied whether ChatGPT can be used for providing natural language explanations for implicit hateful speech detection~\citep{gpt-hate}. Given the promise of zero-shot and few-shot prompting, a series of works~\citep{chatgpt_vs_bert,llm_transform_css} in computer science have compared the performance of prompting GPT models against that of finetuning BERT models and the general consensus is that finetuned BERT models are still overall preferred for text classification despite their much smaller size. In this article, our goal is to situate this question within the field of political science, explore few-shot prompting as a potential solution to data scarcity, and analyze the relative performance of BERT and GPT models in  small-dataset settings.

\section{Empirical Analyses}
\noindent We primarily focus on five sets of experiments: (1) binary classification of news articles based on economic sentiments, (2) 8-class classification of party manifestos,\footnote{Data is collected from mostly democracies in OECD, Central and Eastern European countries and South American countries.} (3) 8-class classification of New Zealand Parliamentary speeches, (4) 20-class classification of COVID-19 policy measures, and (5) 22-class classification of the US State of the Union speeches.

For each experiment, we evaluate the performance of fine-tuning BERT models using 200, 500, and 1,000 samples. The particular BERT version that we use is RoBERTa-large with 340 million parameters~\citep{roberta}.\footnote{Please note that all our experiments utilize RoBERTa-large. For simplicity, we will use the terms BERT and RoBERTa interchangeably in the following sections.} It is arguably the most performant model in the BERT family~\citep{llm_transform_css}. In terms of hyperparameter tuning~\citep{hyperparameters,tutorial,deep_learning}, we optimize the learning rate (3e-5, 2e-5, 1e-5) using the validation set. Each experiment setting is run three times with three different seeds and the mean, min, max are reported.

We further calculate the performance of GPT models with zero sample, 1 sample per class and 2 samples per class, respectively. The particular GPT version we use is GPT-4o.\footnote{Other GPT models include Gemini by Google, Claude by Anthropic, Llama by Meta, Mistral by Mistral AI. The latter two, in particular, offer open-source models. We opt for GPT-4o, which is closed-source, because it arguably provides the best performance. Researchers interested in privacy or latency could consider those smaller open-source alternatives.} In terms of hyperparameter tuning~\citep{zero-shot}, we use two different temperatures: a lower temperature at 0.2, which means less variation in the output, and a higher temperature at 0.8, which means more variation. As in the BERT experiments, each experimental setting is run three times. Each time a particular seed is used for reproducibility. The mean of three runs is reported. For our prompting templates, interested readers could refer to our replication package.

\subsection{Economic Sentiment Classification (2-Class)}
\noindent Sentiment analysis is one of the most common tasks that political scientists have to deal with. Given a particular text snippet, our goal is classify it into either positive or negative. It is often considered an easy task in that it has only two classes. In this experiment, we use the Sentiment Economy News dataset by~\cite{automated_text_analysis} and~\cite{bert_nli}. The goal is to differentiate whether the economy is performing well or poorly according to a given news headline and the corresponding first paragraph~\citep{bert_nli}. In Table~\ref{sentiment}, we report the distribution of the two labels among the train, dev, and test sets. It can be observed that approximately two-thirds of the samples are negative, a pattern consistent across all three datasets. For finetuning the BERT model, we randomly sample 200, 500, and 1,000 samples from the training set. For procuring samples used in few-shot prompting, we randomly select them from the training set as well.

\begin{table}[!h]
\centering
\setlength{\tabcolsep}{20pt}
\caption{Summary statistics of the \textit{Sentiment Economy News} dataset.}\label{sentiment}
\begin{tabular}{llll}
\hline\hline
Dataset    & \multicolumn{1}{c}{Number of samples} & Positive & Negative \\ \hline
Train   & 2000        & 655 (32.75\%)                          &  1345 (67.25\%)         \\
Dev   &  300  &  102  (34.0\%)  & 198 (66.0\%)                                      \\
Test      & 382                                  & 141 (36.91\%)      & 241 (63.09\%) \\ \hline\hline    
\end{tabular}
\end{table}

\begin{figure}[!h]
\centering
\includegraphics[width=418px]{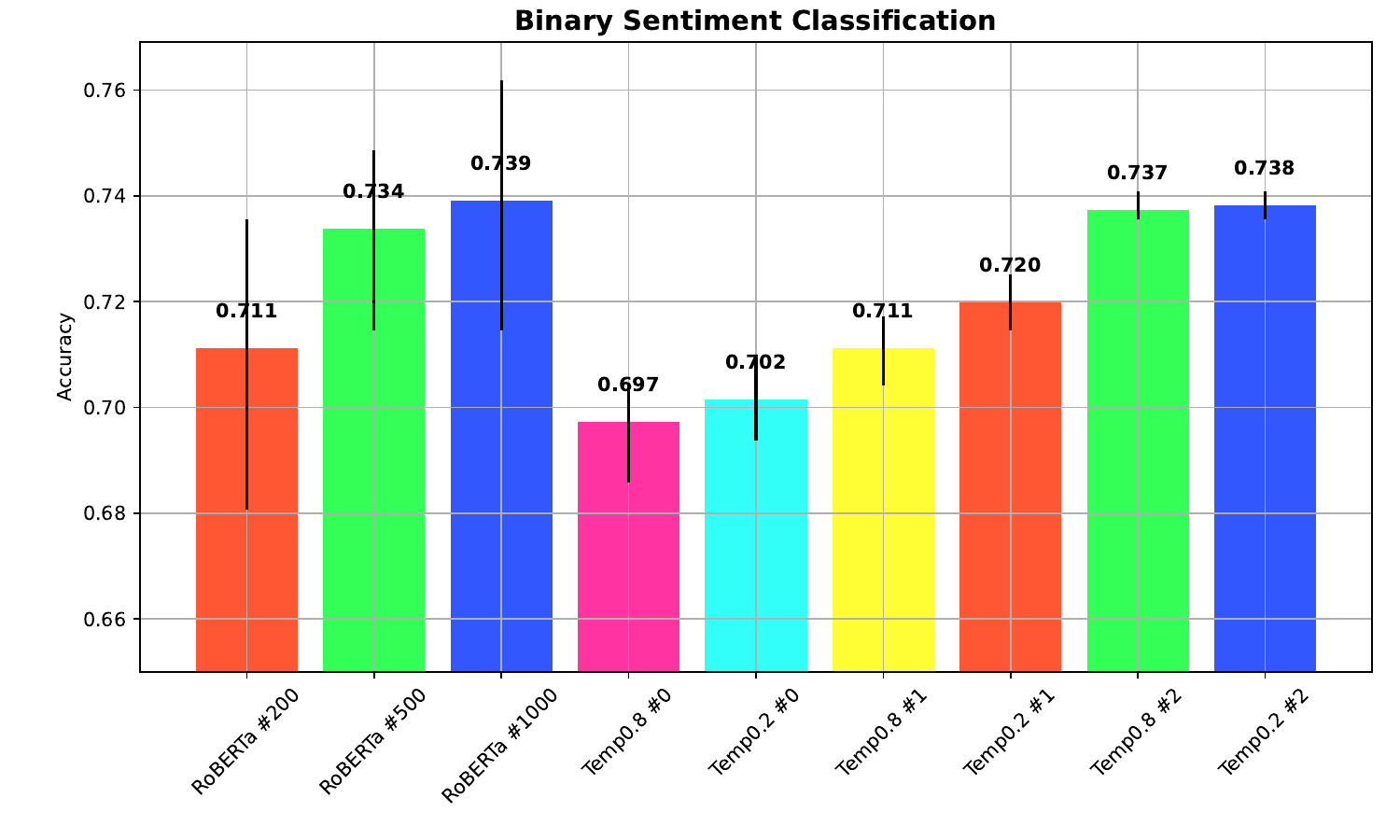}
\caption{Increasing the number of samples enhances model accuracy, whether it's through fine-tuning BERT models or prompting GPT models. `RoBERTa \# 200' refers to fine-tuning RoBERTa-large with 200 samples, while `Temp0.2 \#0' indicates zero-shot prompting with a temperature setting of 0.2. The black vertical error bar represents the range from the minimum to the maximum values. Few-shot prompting with two samples performs about the same as finetuning RoBERTa-large with 1,000 samples. Finetuning yields a higher variance in test evaluations than prompting. A lower temperature setting of 0.2 yields slightly better performance than a higher temperature of 0.8 for prompting.}
\label{fig:sentiment}
\end{figure}

In Figure~\ref{fig:sentiment}, we report the experiment results. We observe that for finetuning BERT models, as the number of training samples increases, the test accuracy increases from 71.1\% (200 samples) to 73.4\% (500 samples) and 73.9\% (1,000 samples). In a similar vein, we observe that one-shot prompting outperforms zero-shot prompting and that two-shot prompting in turns supercedes one-shot prompting. In terms of comparing finetuning BERT models and prompting GPT models, we note that two-shot prompting with ChatGPT matches the performance of finetuning BERT models with 1,000 samples. Zero-shot prompting (70.2\%) is slightly lower than fine-tuning BERT with 200 samples (71.1\%), though the difference is minimal. Additionally, when adjusting the temperature settings in prompting GPT, a temperature of 0.2 offers a slight performance advantage compared to 0.8.

\subsection{Manifesto Classification (8-Class)}
\noindent Topic classification is another common task in political science research~\citep{cross_domain,pretrained_topic_classification}. In terms of the modeling process, it is essentially the same as sentiment analysis, except that it often has more than two classes. In this subsection, we compare the performance of finetuning BERT models with that of prompting GPT models in an 8-class topic classification. The dataset comes from~\cite{bert_nli} and is originally published by WZB Berlin Social Science Center. In this subsection, we further study the problem of 8-class classification of party manifestos.\footnote{Note that data is from the Manifesto Project Dataset and is collected from mostly democracies in OECD, Central and Eastern European countries and South American countries by the WZB Berlin Social Science Center.} In Table~\ref{tab:manifesto_data}, we report the data distribution. The 8 classes are \textit{Economy, External Relations, Fabric of Society, Freedom and Democracy, No Other Category Applies, Political System, Social Groups,} and \textit{Welfare and Quality of Life}. \textit{Economy} and \textit{Welfare and Quality of Life} are the two largest classes, each accounting for between 27\% and 30\%. Other classes are more or less evenly distributed, each accounting for about 9 percent. \textit{No Other Category Applies} is an exception in that it accounts for 0.65\% of the training samples, 1.67\% of the dev samples and 0\% of the test samples. Given how rare this class it, this task effectively boils down to a 7-class classification problem.

\begin{table}[h!]
\centering
\setlength{\tabcolsep}{14pt}
\caption{Distribution in \textit{Manifesto} Datasets.~\textit{Welfare and Quality of Life} alone accounts for nearly a third of the dataset, while~\textit{Economy} represents about one quarter of the dataset. The smallest category, ~\textit{No Other Category Applies}, comprises less than 2\%.}
\begin{tabular}{llll}
\hline\hline
\textbf{Topic} & \textbf{Train} & \textbf{Dev} & \textbf{Test} \\
\hline
Economy & 553 (27.65\%) & 69 (23.0\%) & 85 (28.33\%) \\
External Relations & 141 (7.05\%) & 24 (8.0\%) & 31 (10.33\%) \\
Fabric of Society & 231 (11.55\%) & 34 (11.33\%) & 28 (9.33\%) \\
Freedom and Democracy & 104 (5.2\%) & 23 (7.67\%) & 17 (5.67\%) \\
No Other Category Applies & 13 (0.65\%) & 5 (1.67\%) & - (-) \\
Political System & 180 (9.0\%) & 22 (7.33\%) & 34 (11.33\%) \\
Social Groups & 194 (9.7\%) & 31 (10.33\%) & 22 (7.33\%) \\
Welfare and Quality of Life & 584 (29.2\%) & 92 (30.67\%) & 83 (27.67\%) \\\hline
\textit{Total} & 2000& 300 & 300\\
\hline
\hline
\end{tabular}
\label{tab:manifesto_data}
\end{table}

\begin{figure}[!h]
\centering
\includegraphics[width=408px]{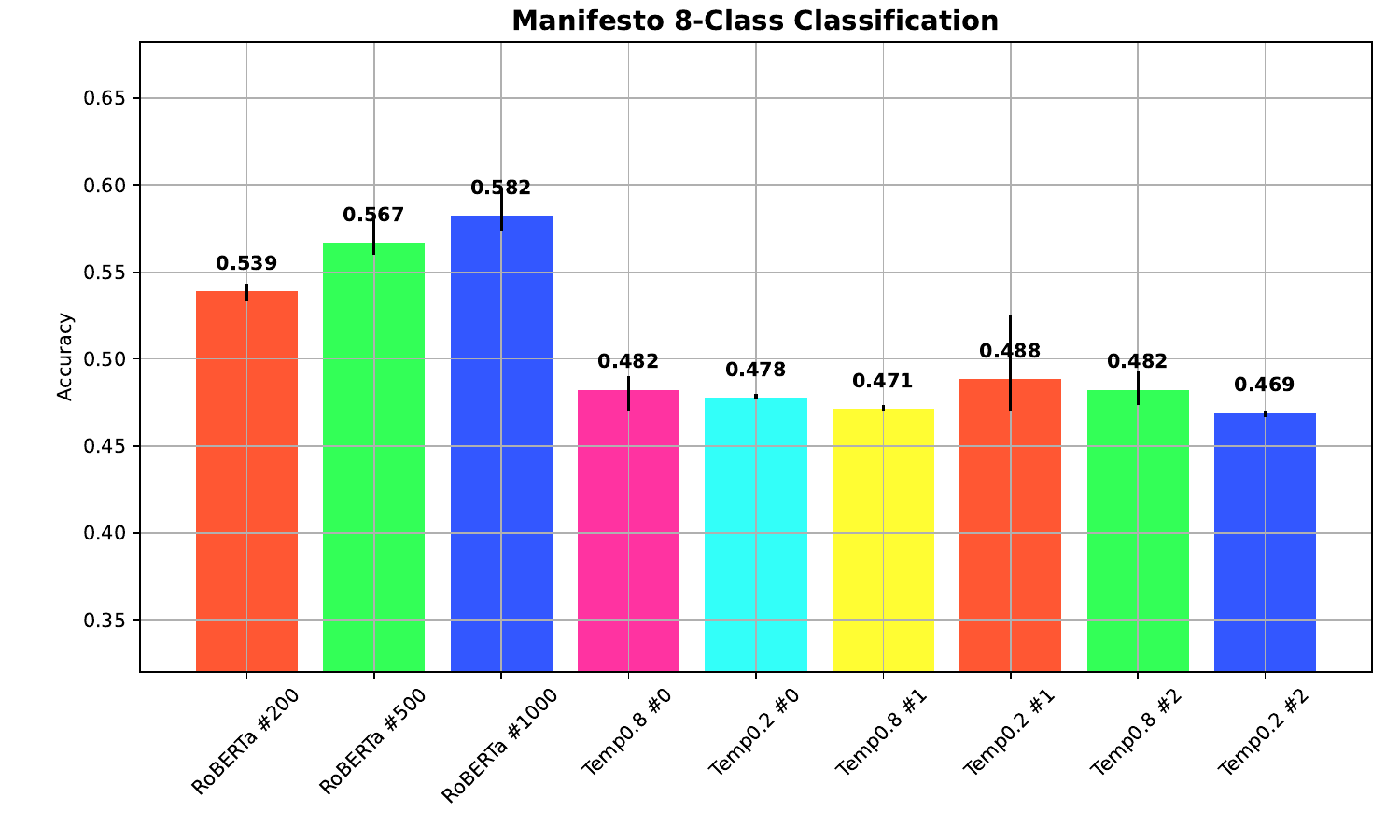}
\caption{Prompting with or without samples lags behind fine-tuning RoBERTa-large models by a sizeable margin in the 8-class manifesto classification.}
\label{manifesto_classification}
\end{figure}

In Figure~\ref{manifesto_classification}, we report our experiment results. We observe that finetuning BERT models yields substantially stronger results than prompting GPT models. As expected, increasing the number of training samples improves the performance of finetuning BERT models. At the same time, zero-shot prompting performs as well as few-shot prompting in this 8-class classification task. When comparing between finetuning BERT models and prompting GPT models, we observe that finetuning with 200 samples yields an accuracy of 53.9\% whereas zero-shot and few-shot prompting yield a maximum accuracy of 48.8\%. While adding extra samples helps further boost the performance of finetuned BERT models, we do not see the same performance gain when adding samples in few-shot prompting.

\subsection{New Zealand Parliamentary Speech Classification (8-Class)}

\noindent In this subsection, we study another example of 8-class classification. Specifically, we classify the speech transcripts from the New Zealand Parliament for the period from 1987 to 2002. The dataset originally comes from~\cite{cross_domain} and has 4,165 hand-coded text snippets. In~\cite{cross_domain}, the authors initially used the dataset as a test set for cross-domain classification.~\cite{pretrained_topic_classification} later split this dataset into train, dev, and test to finetune a BERT model (RoBERT-base). From these 4,165 samples, we random sample 2,000 as the training set, 300 as the dev set, and another 300 as the test set. In Table~\ref{tab:new_zealand_data}, we report the data distribution for our experiment. The data is unbalanced among classes:~\textit{Political System} alone accounts for over a quarter of the dataset, while~\textit{Economy} represents another 17\%. The smallest category, ~\textit{External Relations}, comprises just 2-3\%.

\begin{table}[h!]
\centering
\caption{Distribution in \textit{New Zealand Parliamentary Speech} Datasets.~\textit{Political System} alone accounts for over a quarter of the dataset, while~\textit{Economy} represents another 17\%. The smallest category, ~\textit{External Relations}, comprises just 2-3\%.}
\setlength{\tabcolsep}{14.5pt}
\begin{tabular}{llll}
\hline
\textbf{Topic} & \textbf{Train} & \textbf{Dev} & \textbf{Test} \\
\hline
Economy & 337 (16.85\%) & 57 (19.0\%) & 51 (17.0\%) \\
External Relations & 45 (2.25\%) & 7 (2.33\%) & 8 (2.67\%) \\
Fabric Of Society & 197 (9.85\%) & 30 (10.0\%) & 32 (10.67\%) \\
Freedom and Democracy & 255 (12.75\%) & 34 (11.33\%) & 44 (14.67\%) \\
Other & 100 (5.0\%) & 11 (3.67\%) & 14 (4.67\%) \\
Political System & 541 (27.05\%) & 75 (25.0\%) & 65 (21.67\%) \\
Social Groups & 147 (7.35\%) & 23 (7.67\%) & 26 (8.67\%) \\
Welfare and Quality Of Life & 378 (18.9\%) & 63 (21.0\%) & 60 (20.0\%) \\
\hline
\textit{Total} & 2,000 & 300 & 300\\\hline
\end{tabular}
\label{tab:new_zealand_data}
\end{table}


\begin{figure}[!h]
\centering
\includegraphics[width=408px]{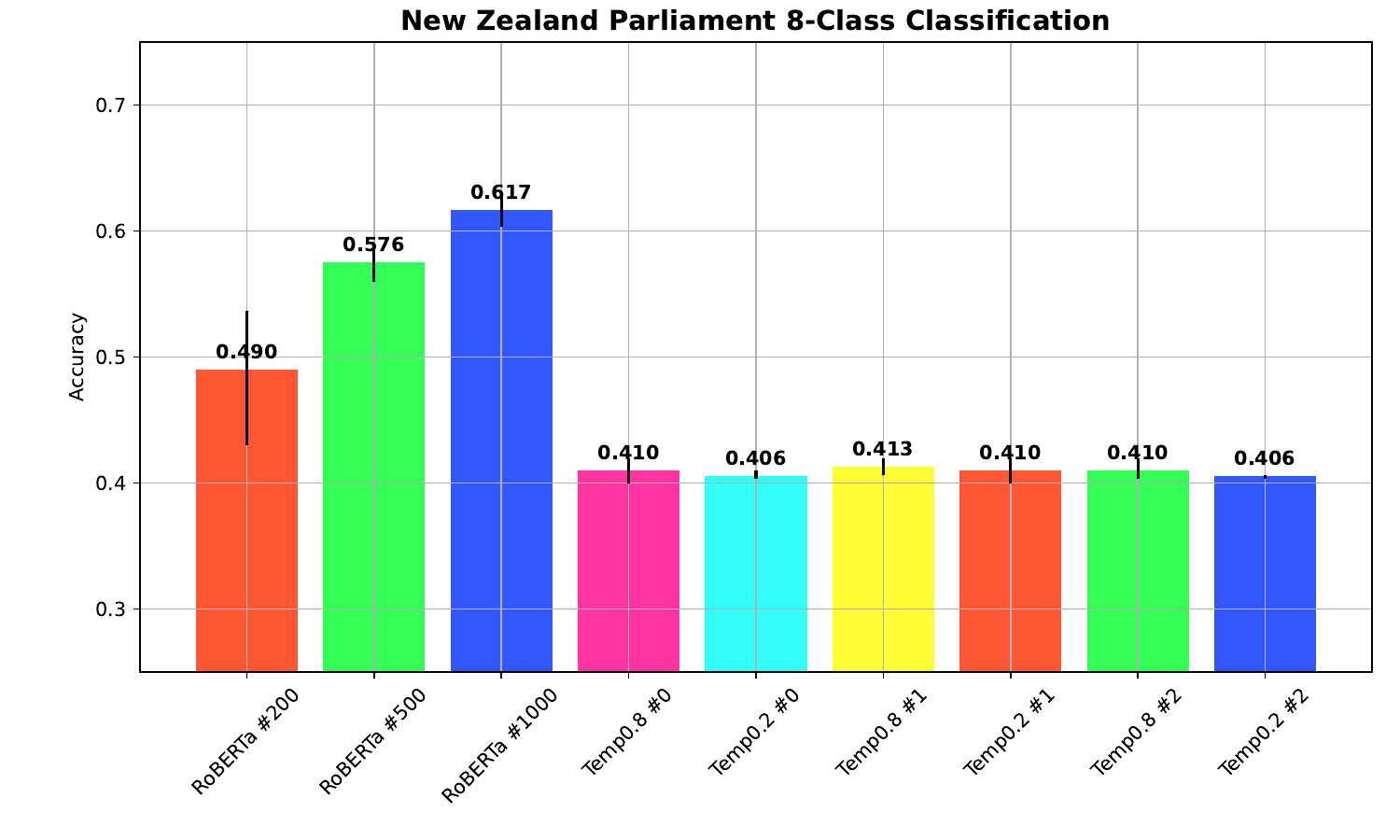}
\caption{Finetuning BERT models substantially outperforms prompting GPT models in the 8-class New Zealand Parliamentary Speech classification. While fine-tuning continues to show significant improvement with the addition of more training samples, prompting appears to gain no benefit from embedding extra samples into the prompts.}
\label{new_zealand_classification}
\end{figure}

In Figure~\ref{new_zealand_classification}, we report our experiment results. A few observations immediately stand out. First, adding more training samples significantly enhances the performance of fine-tuned BERT models. Second, few-shot prompting offers little to no improvement over zero-shot prompting. Third, fine-tuning BERT models is considerably more effective than prompting GPT models. For example, BERT models fine-tuned with 500 samples achieve an accuracy of 57.6\%, nearly 40\% higher than all prompting methods. Furthermore, BERT models fine-tuned with 1,000 samples are about 50\% more accurate than prompting.

\subsection{COVID-19 Policy Measure Classification (20-Class)}

\noindent In this subsection, we evaluate the models' performance on a 20-class classification task. The dataset is in the domain of policy measures against COVID-19. It comes from~\cite{bert_nli} and originally came from~\cite{Cheng2020}. The dataset consists of over 13,000 policy announcements across more than 195 countries and encompasses a total of 20 classes, including, for example, \textit{Curfew} and \textit{External Border Restrictions}. In Table~\ref{tab:corona_data}, we report the data distribution in our experiments. Some of the largest classes, such as \textit{Health Resources} and \textit{Restriction and Regulation of Businesses}, each account for over 10\% of the dataset. In contrast, \textit{Anti-Disinformation Measures} is the smallest class, comprising less than 1\% of the dataset.

\begin{table}[h!]
\centering
\setlength{\tabcolsep}{1pt}
\renewcommand{\arraystretch}{1.1}
\caption{20-class distribution of the COVID-19 Policy Measure dataset. Some of the larger classes include \textit{Health Resources} and~\textit{Restriction and Regulation of Businesses}, each accounting for over 10\% of the dataset.  In contrast,~\textit{Anti-Disinformation Measures} and~\textit{Declaration of Emergency} each account for less than 2\% of the dataset.}
\label{tab:corona_data}
\begin{tabular}{llll}
\hline\hline
\textbf{Topic} & \textbf{Train} & \textbf{Dev} & \textbf{Test} \\
\hline
Anti-Disinformation Measures & 15 (0.75\%) & 0 (0.0\%) & 3 (1.0\%) \\
COVID-19 Vaccines & 66 (3.3\%) & 6 (2.0\%) & 9 (3.0\%) \\
Closure and Regulation of Schools & 114 (5.7\%) & 16 (5.33\%) & 13 (4.33\%) \\
Curfew & 44 (2.2\%) & 6 (2.0\%) & 8 (2.67\%) \\
Declaration of Emergency & 33 (1.65\%) & 6 (2.0\%) & 4 (1.33\%) \\
External Border Restrictions & 119 (5.95\%) & 21 (7.0\%) & 21 (7.0\%) \\
Health Monitoring & 65 (3.25\%) & 12 (4.0\%) & 10 (3.33\%) \\
Health Resources & 283 (14.15\%) & 38 (12.67\%) & 29 (9.67\%) \\
Health Testing & 56 (2.8\%) & 12 (4.0\%) & 9 (3.0\%) \\
Hygiene & 48 (2.4\%) & 4 (1.33\%) & 10 (3.33\%) \\
Internal Border Restrictions & 61 (3.05\%) & 9 (3.0\%) & 12 (4.0\%) \\
Lockdown & 83 (4.15\%) & 10 (3.33\%) & 12 (4.0\%) \\
New Task Force, Bureau or Admin. Configuration & 48 (2.4\%) & 10 (3.33\%) & 9 (3.0\%) \\
Other Policy Not Listed Above & 101 (5.05\%) & 13 (4.33\%) & 21 (7.0\%) \\
Public Awareness Measures & 150 (7.5\%) & 23 (7.67\%) & 21 (7.0\%) \\
Quarantine & 111 (5.55\%) & 21 (7.0\%) & 14 (4.67\%) \\
Restriction and Regulation of Businesses & 227 (11.35\%) & 34 (11.33\%) & 40 (13.33\%) \\
Restriction and Regulation of Govt. Services & 138 (6.9\%) & 17 (5.67\%) & 20 (6.67\%) \\
Restrictions of Mass Gatherings & 139 (6.95\%) & 25 (8.33\%) & 19 (6.33\%) \\
Social Distancing & 99 (4.95\%) & 17 (5.67\%) & 16 (5.33\%) \\
\hline
\textit{Total} & 2000& 300 & 300\\\hline\hline
\end{tabular}
\end{table}

\begin{figure}[!h]
\centering
\includegraphics[width=405px]{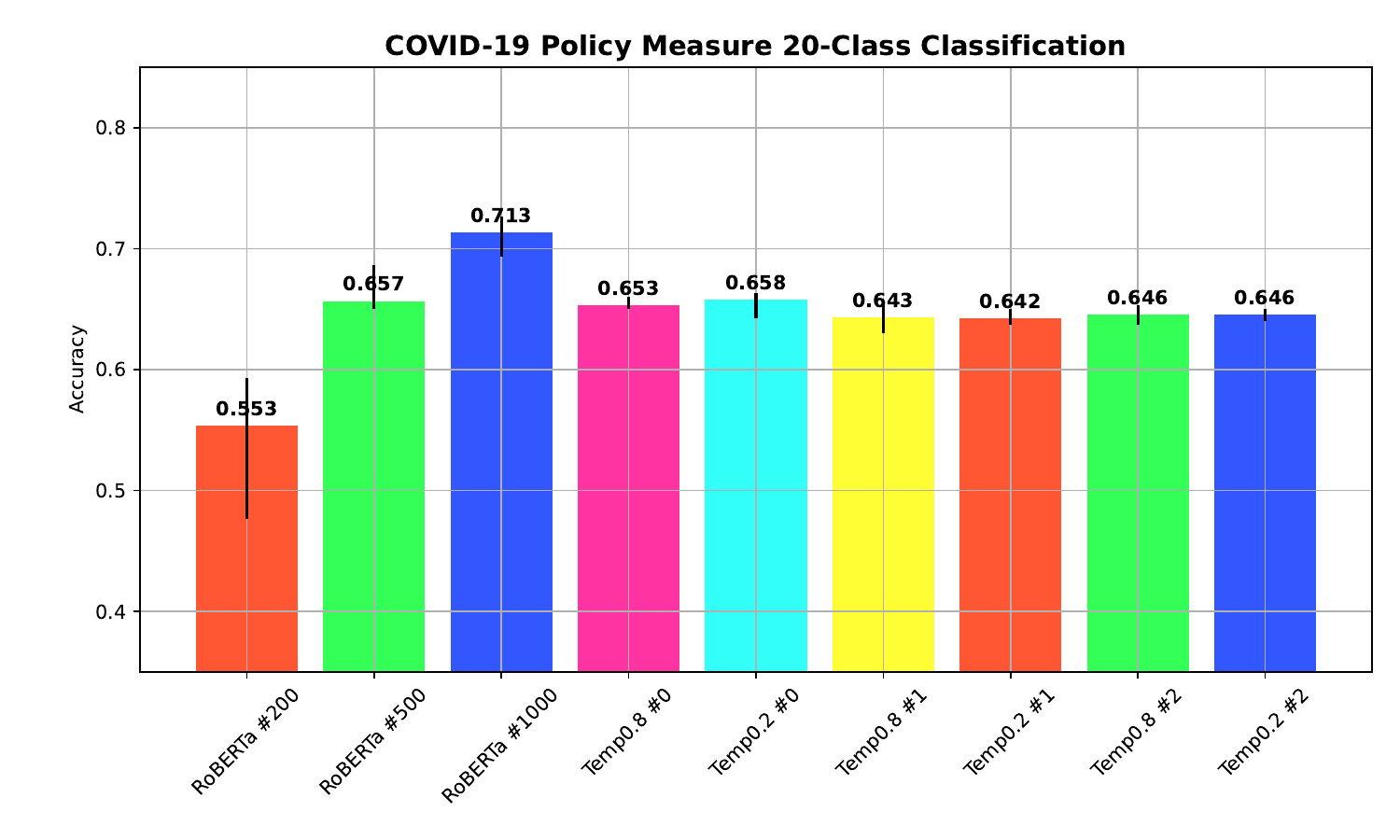}
\caption{Fine-tuning BERT models with 500 samples performs comparably to prompting GPT models in the 20-class COVID-19 policy measure classification task. With fine-tuning continuing to show significant improvement with the addition of more training samples, fine-tuning with 1,000 samples clearly has an edge over prompting, which apparently is not benefiting from the extra added samples.}
\label{corona_classification}
\end{figure}

We report our experiment results in Figure~\ref{corona_classification}. When fine-tuning BERT models, we consistently observe that increasing the number of labeled samples leads to improved performance. When prompting GPT models, however, zero-shot prompting is doing as well as if not better than few-shot prompting. If we compare finetuning and prompting, we observe that prompting GPT models performs at a similar level (65.8\%) as finetuning BERT models with 500 samples (65.7\%). Prompting (65.8\%) is substantially more accurate than finetuning with 200 samples (55.3\%) but is not nearly as good as finetuning with 1,000 samples (71.3\%).

\subsection{Speech Classification (22-Class)}
\noindent Following the 20-class classification task on COVID-19 policy measures, this subsection compares the performance of fine-tuning BERT models with prompting GPT models in a 22-class classification task focused on State of the Union speeches. This task could pose a greater challenge for both approaches, particularly for fine-tuning, for two key reasons. First, with a fixed number of training samples, a larger number of classes means that each class would have fewer samples on average. Second, the fine-tuned BERT model now has significantly more classes to choose from, increasing the likelihood of errors. This second challenge also applies to prompting GPT models. 

\begin{table}[h!]
\centering
\setlength{\tabcolsep}{22.3pt}
\caption{22-class distribution of the State of the Union Speech Dataset. Some of the larger classes are \textit{Defense}, \textit{International Affairs}, and \textit{Macroeconomics}. 
Several topics account for 1\% or less of the dataset, highlighting the uneven distribution of the data.}
\label{tab:sotu_data}
\begin{tabular}{llll}
\hline\hline
\textbf{Topic} & \textbf{Train} & \textbf{Dev} & \textbf{Test} \\
\hline
Agriculture & 36 (2\%) & 9 (3\%) & 6 (2\%) \\
Civil Rights & 47 (2\%) & 6 (2\%) & 7 (2\%) \\
Culture & 0 (0\%) & 0 (0\%) & 1 (0\%) \\
Defense & 281 (14\%) & 38 (13\%) & 34 (11\%) \\
Domestic Commerce & 33 (2\%) & 3 (1\%) & 8 (3\%) \\
Education & 94 (5\%) & 13 (4\%) & 9 (3\%) \\
Energy & 28 (1\%) & 3 (1\%) & 6 (2\%) \\
Environment & 31 (2\%) & 7 (2\%) & 1 (0\%) \\
Foreign Trade & 53 (3\%) & 8 (3\%) & 8 (3\%) \\
Government Operations & 104 (5\%) & 7 (2\%) & 18 (6\%) \\
Health & 79 (4\%) & 11 (4\%) & 11 (4\%) \\
Housing & 30 (1\%) & 2 (1\%) & 7 (2\%) \\
Immigration & 20 (1\%) & 2 (1\%) & 2 (1\%) \\
International Affairs & 282 (14\%) & 52 (17\%) & 37 (12\%) \\
Labor & 52 (3\%) & 8 (3\%) & 23 (8\%) \\
Law and Crime & 67 (3\%) & 13 (4\%) & 11 (4\%) \\
Macroeconomics & 306 (15\%) & 54 (18\%) & 48 (16\%) \\
Other & 318 (16\%) & 42 (14\%) & 51 (17\%) \\
Public Lands & 24 (1\%) & 5 (2\%) & 2 (1\%) \\
Social Welfare & 60 (3\%) & 10 (3\%) & 8 (3\%) \\
Technology & 27 (1\%) & 3 (1\%) & 2 (1\%) \\
Transportation & 28 (1\%) & 4 (1\%) & 0 (0\%) \\
\hline
\textit{Total} & 2000& 300 & 300\\\hline\hline
\end{tabular}
\end{table}

We use the dataset from~\cite{bert_nli}. The data consists of 22 classes: 
\textit{Agriculture}, \textit{Civil Rights}, \textit{Culture}, \textit{Defense}, \textit{Domestic Commerce}, \textit{Education}, \textit{Energy}, \textit{Environment}, \textit{Foreign Trade}, \textit{Government Operations}, \textit{Health}, \textit{Housing}, \textit{Immigration}, \textit{International Affairs}, \textit{Labor}, \textit{Law and Crime}, \textit{Macroeconomics}, \textit{Other}, \textit{Public Lands}, \textit{Social Welfare}, \textit{Technology}, \textit{Transportation}. Some of the larger classes are \textit{Defense} (14\%), \textit{International Affairs} (14\%), and \textit{Macroeconomics} (15\%). Some minor classes, such as \textit{Immigration} and \textit{Technology}, account for 1\% or less of the data (Table~\ref{tab:sotu_data}). As a result, when we sample 200 samples for finetuning BERT models, there are only a couple of samples for these minor classes~\citep{bert_nli}. Quantitatively, that is not dissimilar to the number of samples we use for few-shot prompting.

\begin{figure}[!h]
\centering
\includegraphics[width=408px]{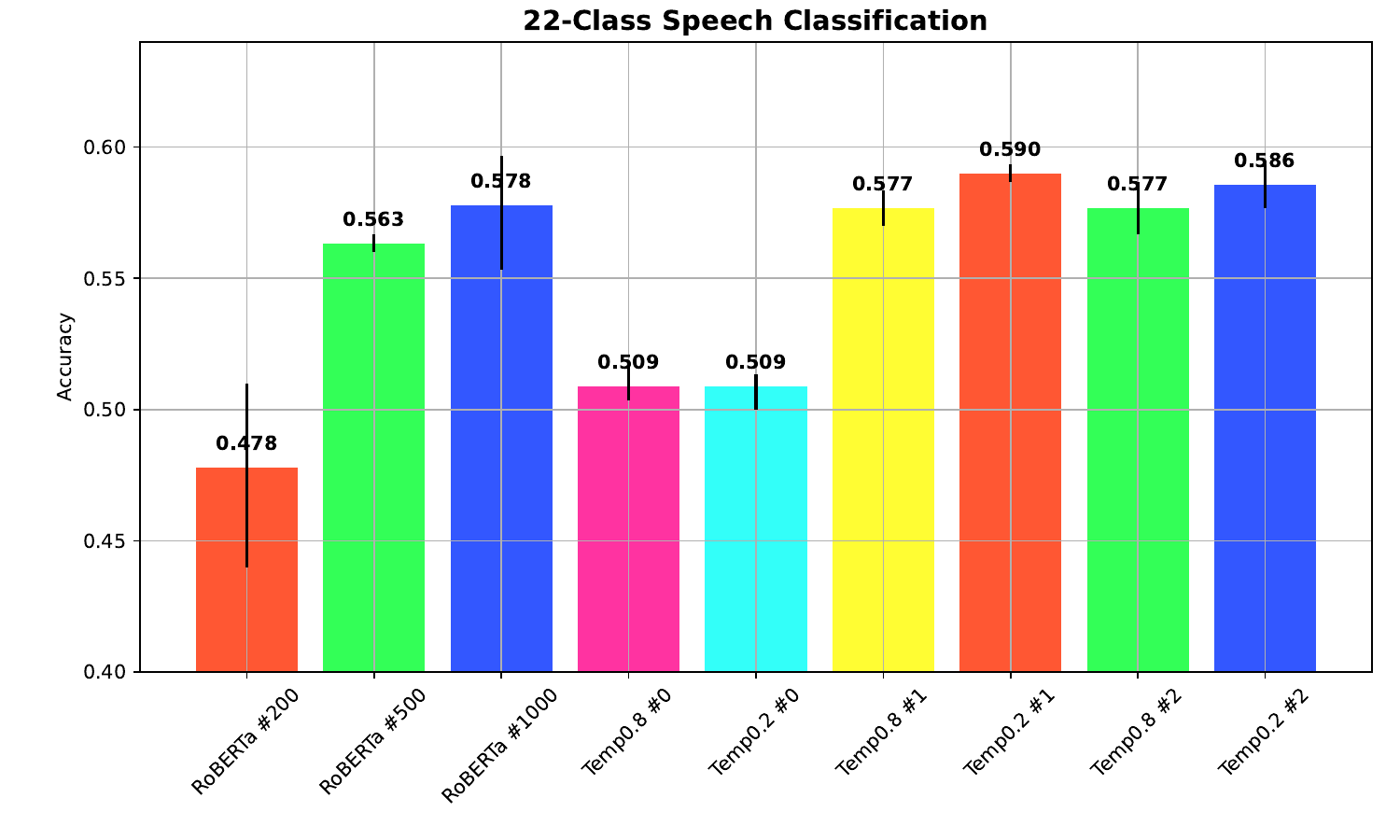}
\caption{In the 22-class classification of the US State of the Union speeches, zero-shot prompting outperforms finetuning BERT with 200 samples. 1-shot and 2-shot prompting perform similarly to finetuning with 500 and 1,000 samples, respectively.}
\label{speech_classification}
\end{figure}

We report our experiment results in Figure~\ref{speech_classification}. Regarding finetuning BERT models, our typical observation holds true: increasing the number of training samples from 200 to 500 and then to 1,000 consistently results in performance improvements. When prompting GPT models, we observe that while the difference in performance between 1-shot and 2-shot prompting is minimal, the improvement from zero-shot to few-shot prompting is significant, with accuracy increasing from 0.509 to 0.590. Between finetuning BERT and prompting GPT, we note that zero-shot prompting outperforms finetuning with 200 samples and that few-shot prompting is equal to or slightly better than finetuning BERT models with 500 or 1,000 samples.

\section{Discussions and Future Research}
\noindent The empirical results consistently demonstrate that fine-tuning BERT models is the preferred approach for maximizing model accuracy when researchers have access to around 1,000 data points. However, while prompting may not achieve the same level of performance, it offers competitive results, particularly when the training set includes only a few hundred samples. In this section, we delve deeper into these findings, discussing them in terms of performance, ease of use, cost considerations, and potential future directions.


\subsection{Performance}
\noindent After comparing the performance of fine-tuning BERT models and prompting GPT models across binary, 8-class, and 20+ class classifications, several key observations immediately stand out. First and foremost, in general both finetuning and prompting represent viable solutions to the data scarcity issue and both offer strong performance with limited labeled data. Second, when comparing the performance of these two approaches, we note that for certain tasks, binary classifications in particular, zero-shot and few-shot prompting can already do as well as BERT finetuning. Third, if the goal is model accuracy, then researchers will have better success with finetuning. In Table~\ref{performance}, we summarize the comparisons between finetuning BERT and prompting GPT in terms of the required amount of data and optimal task difficulty. From a practical standpoint, since political scientists often use classification results as input for regression analysis~\citep{unsupervised_semi_supervised_visual_frames,fong2021machine}, both approaches enable researchers to quickly start experimenting with new ideas, even with minimal or no labeled data~\citep{bert_nli,pretrained_topic_classification,aug}. This is especially true for zero-shot and few-shot prompting, which we will explore next.

\begin{table}[!h]
\caption{Selection of BERT and GPT Models Based on Data Availability and Task Difficulty.}
\setlength{\tabcolsep}{18pt}
\label{performance}
\begin{tabular}{lll}
\hline\hline
                     & Required Data Amount & Optimal Task Difficulty \\\hline
Fine-tuning BERT     & Small                & High                    \\
Prompting GPT        & Minimal to None      & Low          \\\hline \hline           
\end{tabular}
\end{table}



\subsection{Ease of Use}
\noindent In terms of ease to use, finetuning BERT models is more complicated than zero-shot or few-shot prompting GPT models. In terms of data preparation, both approaches represent an advancement over classical methods, since there is no more need for data preprocessing, such as stop word removal and stemming~\citep{bag_of_words}. For finetuning BERT models, we need to split the dataset into train, dev, and test. For prompting, we only need the test set (and an optional dev set). When it comes to training, fine-tuning has been greatly simplified by frameworks like Huggingface~\citep{bert_nli}. However, researchers still need to write some boilerplate code. Additionally, there is the need to adjust quite a few parameters, with the learning rate being particularly important~\citep{hyperparameters,deep_learning}. By contrast, prompting with GPT models is done via API calls and requires little code. Temperature is arguably the only hyperparameter that researchers need to deal with~\citep{zero-shot}.

\begin{table}[!h]
\caption{Comparing Finetuning BERT Models and Prompting GPT Models in Terms of Ease of Use}
\centering
\setlength{\tabcolsep}{21pt}
\begin{tabular}{llll}
\hline\hline
                 & Data Processing & Training  & Evaluation \\\hline
Finetuning BERT  & Medium           & Medium    & Easy       \\
Prompting GPT    & Easy             & None      & Easy      \\\hline
\end{tabular}
\end{table}

\subsection{Cost}
\noindent In addition to performance and ease of use, a third dimension to consider is the financial cost of using these models.\footnote{Researchers may also need to consider the cost of annotation, which is often not trivial~\citep{zero-shot}.} The cost of fine-tuning BERT models primarily lies in GPU time: the time spent using GPUs for fine-tuning and evaluation/inference. As we increase the number of training samples from 200 to 500 and then to 1,000, we will linearly increase the training time and thus the cost. As an example, in the 20-class classification of COVID-19 policy measures, training a BERT model with 200 samples takes three and a half minute. For 500 samples, it takes five and a half minute. For 1000 samples, it takes eight and a half minute. During evaluation (inference), each sample takes about 10 milliseconds. As we increase the number of test samples, we linearly increase the cost. Since fine-tuning is performed only once, the associated costs can be considered sunk. In contrast, prompting does not involve fine-tuning, so there are no sunk costs. However, each individual API call for prompting is typically more expensive than the cost of BERT inferencing, at least for now. Additionally, the cost of prompting GPT models is tied to the number of tokens in the prompt—the more tokens per request, the higher the cost.









\begin{figure}[!h]
\centering
\includegraphics[width=358px]{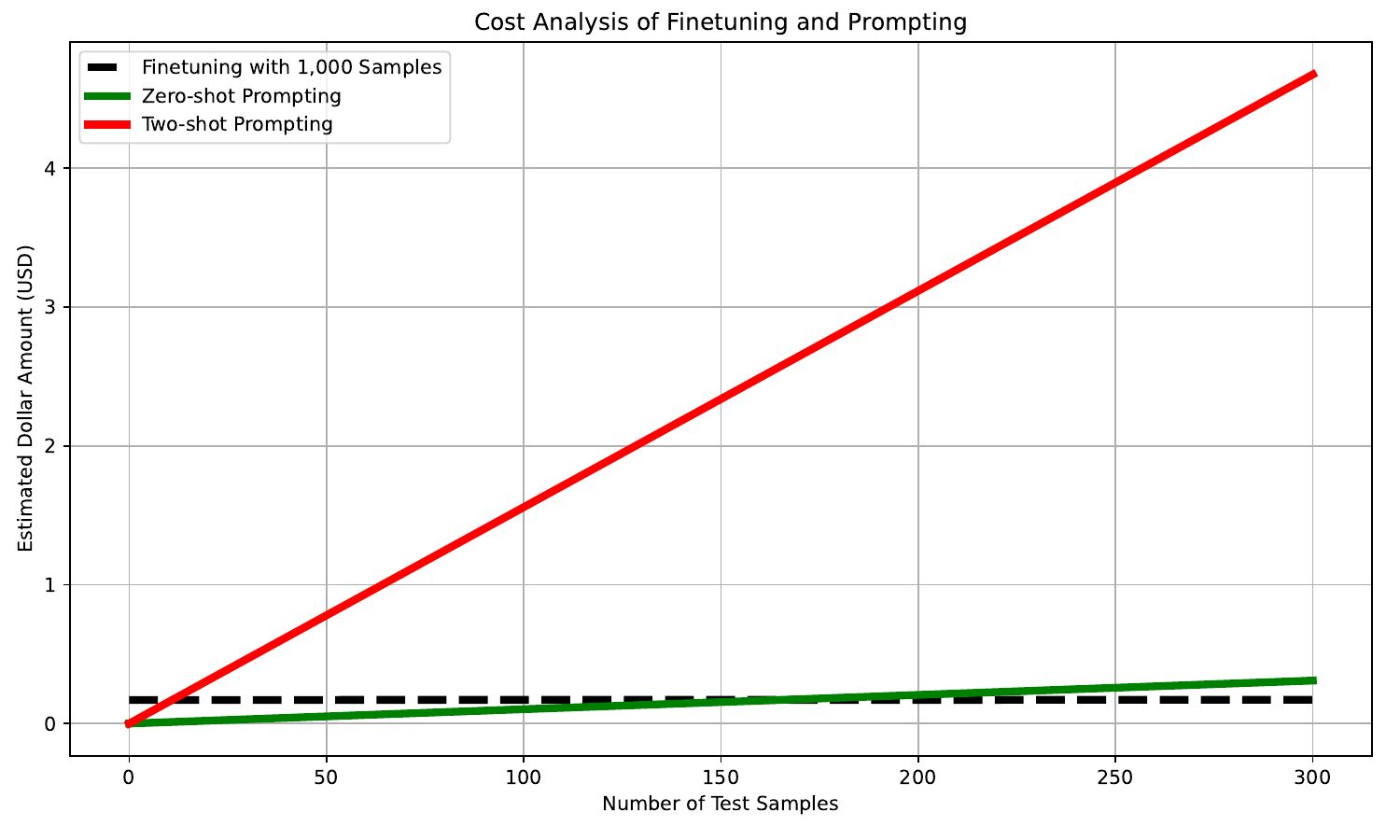}
\caption{When fine-tuning BERT models, there is a sunk cost associated with the initial fine-tuning process. However, this approach proves to be more economical during inference. In contrast, prompting incurs no such sunk cost but has a steeper cost curve for inference. In this experiment, the cost of zero-shot prompting catches up with fine-tuning after processing 150-200 samples, while the cost of 2-shot prompting quickly surpasses that of fine-tuning after just a few API calls.}
\label{cost_analysis}
\end{figure}

In Figure~\ref{cost_analysis}, we compare the cost of finetuning BERT models with that of prompting GPT models. Our comparison is based on the following assumptions: (1) the cost of running an A100 GPU is  \$1.20 per hour, and (2) the cost of prompting GPT-4o is \$5 per million tokens.\footnote{For more details on fine-tuning and GPU costs, please visit https://colab.research.google.com/signup. For prompting and ChatGPT pricing, please see https://openai.com/api/pricing/.} The black dashed line represents the cost of using fine-tuned BERT models. While there is an initial sunk cost associated with fine-tuning on 1,000 samples, the resulting model demonstrates a relatively flat cost slope. During inference, the fine-tuned model incurs additional computational costs at a rate of 100 samples per second. For zero-shot prompting (green solid line), there is no initial cost, but it has a steeper slope, intersecting with the fine-tuning cost curve at 150-200 test samples. Two-shot prompting is the most expensive approach. In this policy measure classification task with 20 classes, each prompt includes 40 additional examples (two for each class). Consequently, the slope is significantly steeper compared to zero-shot prompting. For a more detailed comparison, interested readers may consider exploring other GPUs, such as the H100, and alternative generative AI models such as Gemini Pro and GPT-4 mini.


\subsection{Future Directions}
\noindent Natural language processing (NLP) and large language models are advancing rapidly. In this section, we outline several emerging directions in NLP that hold significant promise for enhancing political science research. Of the two approaches, fine-tuning BERT models is more mature, while prompting is still relatively new. For fine-tuning BERT models, researchers could further explore mixed precision training to reduce the sunk cost, particularly when working with large datasets. To enhance the performance of GPT models, researchers might investigate newer foundation models, advanced prompting techniques such as chain-of-thought and self-consistency~\citep{wei2022chain,wang2023self}, and more effective sample selection methods based on criteria like semantic similarity~\citep{liu2022what,an2023skill}. These strategies could not only improve the effectiveness of prompting GPT models but also make them more economically compelling.



\section{Conclusion}
\noindent Quantitative text analysis plays a prominent role in political science research. Recent advancements, particularly in large language models, have provided researchers with powerful new tools to address both existing and emerging challenges. In this article, we have explored the potential of using GPT-based models as an alternative solution to the data scarcity issue, comparing their performance to that of fine-tuning BERT models, which remains the state of the art. Through extensive experiments, we have demonstrated that zero-shot and few-shot learning with GPT-based models can sometimes serve as an effective alternative to fine-tuning BERT models, especially when the number of classes is small, but fine-tuning BERT models remains the overall go-to method for classification. In addition to performance, we have also compared these approaches in terms of ease of use and cost. While prompting GPT models is significantly easier to use than fine-tuning BERT models, it also proves to be more expensive. We believe our findings will be valuable to researchers involved in quantitative text analysis.

\section*{Data Availability}
\noindent All our data and code will be made publicly available and posted on Harvard Dataverse upon the paper's acceptance.

\section*{Competing interests}
\noindent The author(s) declare no competing interests.


\section*{Ethical approval}
\noindent This article does not contain any studies with human participants performed by any of the authors.

\section*{Informed consent}
\noindent This article does not contain any studies with human participants performed by any of the authors.


\bibliographystyle{apacite}
\bibliography{ir}

\begin{thebibliography}{}

\bibitem [\protect \citeauthoryear {%
Adams%
\ \protect \BOthers {.}}{%
Adams%
\ \protect \BOthers {.}}{%
{\protect \APACyear {2023}}%
}]{%
gpt4_radiology_reports_transformation}
\APACinsertmetastar {%
gpt4_radiology_reports_transformation}%
\begin{APACrefauthors}%
Adams, L\BPBI C.%
, Truhn, D.%
, Busch, F.%
, Kader, A.%
, Niehues, S\BPBI M.%
, Makowski, M\BPBI R.%
\BCBL {}\ \BBA {} Bressem, K\BPBI K.%
\end{APACrefauthors}%
\unskip\
\newblock
\APACrefYearMonthDay{2023}{}{}.
\newblock
{\BBOQ}\APACrefatitle {Leveraging GPT-4 for Post Hoc Transformation of Free-text Radiology Reports into Structured Reporting: A Multilingual Feasibility Study} {Leveraging gpt-4 for post hoc transformation of free-text radiology reports into structured reporting: A multilingual feasibility study}.{\BBCQ}
\newblock
\APACjournalVolNumPages{Radiology}{}{}{}.
\newblock
\begin{APACrefDOI} \doi{https://doi.org/10.1148/radiol.230725} \end{APACrefDOI}
\PrintBackRefs{\CurrentBib}

\bibitem [\protect \citeauthoryear {%
An%
\ \protect \BOthers {.}}{%
An%
\ \protect \BOthers {.}}{%
{\protect \APACyear {2023}}%
}]{%
an2023skill}
\APACinsertmetastar {%
an2023skill}%
\begin{APACrefauthors}%
An, S.%
, Zhou, B.%
, Lin, Z.%
, Fu, Q.%
, Chen, B.%
, Zheng, N.%
\BDBL {}Lou, J\BHBI G.%
\end{APACrefauthors}%
\unskip\
\newblock
\APACrefYearMonthDay{2023}{}{}.
\newblock
{\BBOQ}\APACrefatitle {Skill-Based Few-Shot Selection for In-Context Learning} {Skill-based few-shot selection for in-context learning}.{\BBCQ}
\newblock
\BIn{} \APACrefbtitle {Proceedings of the 2023 Conference on Empirical Methods in Natural Language Processing (EMNLP).} {Proceedings of the 2023 conference on empirical methods in natural language processing (emnlp).}
\PrintBackRefs{\CurrentBib}

\bibitem [\protect \citeauthoryear {%
Argyle%
, Bail%
\BCBL {}\ \protect \BOthers {.}}{%
Argyle%
, Bail%
\BCBL {}\ \protect \BOthers {.}}{%
{\protect \APACyear {2023}}%
}]{%
democratic_discourse}
\APACinsertmetastar {%
democratic_discourse}%
\begin{APACrefauthors}%
Argyle, L\BPBI P.%
, Bail, C\BPBI A.%
, Busby, E\BPBI C.%
, Gubler, J\BPBI R.%
, Howe, T.%
, Rytting, C.%
\BDBL {}Wingate, D.%
\end{APACrefauthors}%
\unskip\
\newblock
\APACrefYearMonthDay{2023}{}{}.
\newblock
{\BBOQ}\APACrefatitle {Leveraging AI for Democratic Discourse: Chat Interventions can Improve Online Political Conversations at Scale} {Leveraging ai for democratic discourse: Chat interventions can improve online political conversations at scale}.{\BBCQ}
\newblock
\APACjournalVolNumPages{Proceedings of the National Academy of Sciences}{}{}{}.
\PrintBackRefs{\CurrentBib}

\bibitem [\protect \citeauthoryear {%
Argyle%
, Busby%
\BCBL {}\ \protect \BOthers {.}}{%
Argyle%
, Busby%
\BCBL {}\ \protect \BOthers {.}}{%
{\protect \APACyear {2023}}%
}]{%
argyle_busby_fulda_gubler_rytting_wingate_2023}
\APACinsertmetastar {%
argyle_busby_fulda_gubler_rytting_wingate_2023}%
\begin{APACrefauthors}%
Argyle, L\BPBI P.%
, Busby, E\BPBI C.%
, Fulda, N.%
, Gubler, J\BPBI R.%
, Rytting, C.%
\BCBL {}\ \BBA {} Wingate, D.%
\end{APACrefauthors}%
\unskip\
\newblock
\APACrefYearMonthDay{2023}{}{}.
\newblock
{\BBOQ}\APACrefatitle {Out of One, Many: Using Language Models to Simulate Human Samples} {Out of one, many: Using language models to simulate human samples}.{\BBCQ}
\newblock
\APACjournalVolNumPages{Political Analysis}{}{}{1–15}.
\newblock
\begin{APACrefDOI} \doi{10.1017/pan.2023.2} \end{APACrefDOI}
\PrintBackRefs{\CurrentBib}

\bibitem [\protect \citeauthoryear {%
Arnold%
, Biedebach%
, Küpfer%
\BCBL {}\ \BBA {} Neunhoeffer%
}{%
Arnold%
\ \protect \BOthers {.}}{%
{\protect \APACyear {2024}}%
}]{%
hyperparameters}
\APACinsertmetastar {%
hyperparameters}%
\begin{APACrefauthors}%
Arnold, C.%
, Biedebach, L.%
, Küpfer, A.%
\BCBL {}\ \BBA {} Neunhoeffer, M.%
\end{APACrefauthors}%
\unskip\
\newblock
\APACrefYearMonthDay{2024}{}{}.
\newblock
{\BBOQ}\APACrefatitle {The Role of Hyperparameters in Machine Learning Models and How to Tune Them} {The role of hyperparameters in machine learning models and how to tune them}.{\BBCQ}
\newblock
\APACjournalVolNumPages{Political Science Research and Methods}{}{}{}.
\PrintBackRefs{\CurrentBib}

\bibitem [\protect \citeauthoryear {%
Barberá%
, Boydstun%
, Linn%
, McMahon%
\BCBL {}\ \BBA {} Nagler%
}{%
Barberá%
\ \protect \BOthers {.}}{%
{\protect \APACyear {2021}}%
}]{%
automated_text_analysis}
\APACinsertmetastar {%
automated_text_analysis}%
\begin{APACrefauthors}%
Barberá, P.%
, Boydstun, A\BPBI E.%
, Linn, S.%
, McMahon, R.%
\BCBL {}\ \BBA {} Nagler, J.%
\end{APACrefauthors}%
\unskip\
\newblock
\APACrefYearMonthDay{2021}{}{}.
\newblock
{\BBOQ}\APACrefatitle {Automated Text Classification Of News Articles: A Practical Guide} {Automated text classification of news articles: A practical guide}.{\BBCQ}
\newblock
\APACjournalVolNumPages{Political Anslysis}{}{}{}.
\PrintBackRefs{\CurrentBib}

\bibitem [\protect \citeauthoryear {%
Bestvater%
\ \BBA {} Monroe%
}{%
Bestvater%
\ \BBA {} Monroe%
}{%
{\protect \APACyear {2023}}%
}]{%
bestvater_monroe_2023}
\APACinsertmetastar {%
bestvater_monroe_2023}%
\begin{APACrefauthors}%
Bestvater, S\BPBI E.%
\BCBT {}\ \BBA {} Monroe, B\BPBI L.%
\end{APACrefauthors}%
\unskip\
\newblock
\APACrefYearMonthDay{2023}{}{}.
\newblock
{\BBOQ}\APACrefatitle {Sentiment is Not Stance: Target-Aware Opinion Classification for Political Text Analysis} {Sentiment is not stance: Target-aware opinion classification for political text analysis}.{\BBCQ}
\newblock
\APACjournalVolNumPages{Political Analysis}{31}{2}{235–256}.
\newblock
\begin{APACrefDOI} \doi{10.1017/pan.2022.10} \end{APACrefDOI}
\PrintBackRefs{\CurrentBib}

\bibitem [\protect \citeauthoryear {%
Bisbee%
, Clinton%
, Dorff%
, Kenkel%
\BCBL {}\ \BBA {} Larson%
}{%
Bisbee%
\ \protect \BOthers {.}}{%
{\protect \APACyear {2024}}%
}]{%
synthetic_replacement}
\APACinsertmetastar {%
synthetic_replacement}%
\begin{APACrefauthors}%
Bisbee, J.%
, Clinton, J\BPBI D.%
, Dorff, C.%
, Kenkel, B.%
\BCBL {}\ \BBA {} Larson, J\BPBI M.%
\end{APACrefauthors}%
\unskip\
\newblock
\APACrefYearMonthDay{2024}{}{}.
\newblock
{\BBOQ}\APACrefatitle {Synthetic Replacements for Human Survey Data? The Perils of Large Language Models} {Synthetic replacements for human survey data? the perils of large language models}.{\BBCQ}
\newblock
\APACjournalVolNumPages{Political Analysis}{}{}{}.
\PrintBackRefs{\CurrentBib}

\bibitem [\protect \citeauthoryear {%
Brown%
\ \protect \BOthers {.}}{%
Brown%
\ \protect \BOthers {.}}{%
{\protect \APACyear {2020}}%
}]{%
gpt3}
\APACinsertmetastar {%
gpt3}%
\begin{APACrefauthors}%
Brown, T.%
, Mann, B.%
, Ryder, N.%
, Subbiah, M.%
, Kaplan, J\BPBI D.%
, Dhariwal, P.%
\BDBL {}Amodei, D.%
\end{APACrefauthors}%
\unskip\
\newblock
\APACrefYearMonthDay{2020}{}{}.
\newblock
{\BBOQ}\APACrefatitle {Language Models are Few-Shot Learners} {Language models are few-shot learners}.{\BBCQ}
\newblock
\APACjournalVolNumPages{34th Conference on Neural Information Processing Systems (NeurIPS 2020)}{}{}{}.
\PrintBackRefs{\CurrentBib}

\bibitem [\protect \citeauthoryear {%
Chang%
\ \BBA {} Masterson%
}{%
Chang%
\ \BBA {} Masterson%
}{%
{\protect \APACyear {2020}}%
}]{%
lstm_pa}
\APACinsertmetastar {%
lstm_pa}%
\begin{APACrefauthors}%
Chang, C.%
\BCBT {}\ \BBA {} Masterson, M.%
\end{APACrefauthors}%
\unskip\
\newblock
\APACrefYearMonthDay{2020}{}{}.
\newblock
{\BBOQ}\APACrefatitle {Using Word Order in Political Text Classification With Long Short-Term Memory Models} {Using word order in political text classification with long short-term memory models}.{\BBCQ}
\newblock
\APACjournalVolNumPages{Political Analysis}{}{}{}.
\PrintBackRefs{\CurrentBib}

\bibitem [\protect \citeauthoryear {%
Chaturvedi%
\ \BBA {} Chaturvedi%
}{%
Chaturvedi%
\ \BBA {} Chaturvedi%
}{%
{\protect \APACyear {2023}}%
}]{%
all_in_the_name}
\APACinsertmetastar {%
all_in_the_name}%
\begin{APACrefauthors}%
Chaturvedi, R.%
\BCBT {}\ \BBA {} Chaturvedi, S.%
\end{APACrefauthors}%
\unskip\
\newblock
\APACrefYearMonthDay{2023}{}{}.
\newblock
{\BBOQ}\APACrefatitle {It's All in the Name: A Character-Based Approach to Infer Religion} {It's all in the name: A character-based approach to infer religion}.{\BBCQ}
\newblock
\APACjournalVolNumPages{Political Analysis}{}{}{}.
\PrintBackRefs{\CurrentBib}

\bibitem [\protect \citeauthoryear {%
Cheng%
, Barceló%
, Hartnett%
, Kubinec%
\BCBL {}\ \BBA {} Messerschmidt%
}{%
Cheng%
\ \protect \BOthers {.}}{%
{\protect \APACyear {2020}}%
}]{%
Cheng2020}
\APACinsertmetastar {%
Cheng2020}%
\begin{APACrefauthors}%
Cheng, C.%
, Barceló, J.%
, Hartnett, A\BPBI S.%
, Kubinec, R.%
\BCBL {}\ \BBA {} Messerschmidt, L.%
\end{APACrefauthors}%
\unskip\
\newblock
\APACrefYearMonthDay{2020}{}{}.
\newblock
{\BBOQ}\APACrefatitle {{COVID-19 Government Response Event Dataset (CoronaNet v.1.0)}} {{COVID-19 Government Response Event Dataset (CoronaNet v.1.0)}}.{\BBCQ}
\newblock
\APACjournalVolNumPages{Nature Human Behaviour}{4}{}{756--768}.
\newblock
\begin{APACrefDOI} \doi{10.1038/s41562-020-0909-7} \end{APACrefDOI}
\PrintBackRefs{\CurrentBib}

\bibitem [\protect \citeauthoryear {%
Devlin%
, Chang%
, Lee%
\BCBL {}\ \BBA {} Toutanova%
}{%
Devlin%
\ \protect \BOthers {.}}{%
{\protect \APACyear {2019}}%
}]{%
bert}
\APACinsertmetastar {%
bert}%
\begin{APACrefauthors}%
Devlin, J.%
, Chang, M\BHBI W.%
, Lee, K.%
\BCBL {}\ \BBA {} Toutanova, K.%
\end{APACrefauthors}%
\unskip\
\newblock
\APACrefYearMonthDay{2019}{}{}.
\newblock
{\BBOQ}\APACrefatitle {BERT: Pre-training of Deep Bidirectional Transformers for Language Understanding} {Bert: Pre-training of deep bidirectional transformers for language understanding}.{\BBCQ}
\newblock
\APACjournalVolNumPages{Proceedings of NAACL-HLT}{}{}{4171-4186}.
\PrintBackRefs{\CurrentBib}

\bibitem [\protect \citeauthoryear {%
Diermeier%
, Godbout%
, Yu%
\BCBL {}\ \BBA {} Kaufmann%
}{%
Diermeier%
\ \protect \BOthers {.}}{%
{\protect \APACyear {2011}}%
}]{%
language_ideology}
\APACinsertmetastar {%
language_ideology}%
\begin{APACrefauthors}%
Diermeier, D.%
, Godbout, J\BHBI F.%
, Yu, B.%
\BCBL {}\ \BBA {} Kaufmann, S.%
\end{APACrefauthors}%
\unskip\
\newblock
\APACrefYearMonthDay{2011}{}{}.
\newblock
{\BBOQ}\APACrefatitle {Language and Ideology in Congress} {Language and ideology in congress}.{\BBCQ}
\newblock
\APACjournalVolNumPages{British Journal of Political Science}{}{}{}.
\PrintBackRefs{\CurrentBib}

\bibitem [\protect \citeauthoryear {%
D'Orazio%
, Landis%
, Palmer%
\BCBL {}\ \BBA {} Schrodt%
}{%
D'Orazio%
\ \protect \BOthers {.}}{%
{\protect \APACyear {2014}}%
}]{%
wheat_chaff}
\APACinsertmetastar {%
wheat_chaff}%
\begin{APACrefauthors}%
D'Orazio, V.%
, Landis, S\BPBI T.%
, Palmer, G.%
\BCBL {}\ \BBA {} Schrodt, P.%
\end{APACrefauthors}%
\unskip\
\newblock
\APACrefYearMonthDay{2014}{}{}.
\newblock
{\BBOQ}\APACrefatitle {Separating the Wheat from the Chaff: Applications of Automated Document Classification Using Support Vector Machines} {Separating the wheat from the chaff: Applications of automated document classification using support vector machines}.{\BBCQ}
\newblock
\APACjournalVolNumPages{Political Anslysis}{}{}{}.
\PrintBackRefs{\CurrentBib}

\bibitem [\protect \citeauthoryear {%
Fong%
\ \BBA {} Tyler%
}{%
Fong%
\ \BBA {} Tyler%
}{%
{\protect \APACyear {2021}}%
}]{%
fong2021machine}
\APACinsertmetastar {%
fong2021machine}%
\begin{APACrefauthors}%
Fong, C.%
\BCBT {}\ \BBA {} Tyler, M.%
\end{APACrefauthors}%
\unskip\
\newblock
\APACrefYearMonthDay{2021}{October}{}.
\newblock
{\BBOQ}\APACrefatitle {Machine Learning Predictions as Regression Covariates} {Machine learning predictions as regression covariates}.{\BBCQ}
\newblock
\APACjournalVolNumPages{Political Analysis}{29}{4}{467--484}.
\PrintBackRefs{\CurrentBib}

\bibitem [\protect \citeauthoryear {%
Gilardi%
, Alizadeh%
\BCBL {}\ \BBA {} Kubli%
}{%
Gilardi%
\ \protect \BOthers {.}}{%
{\protect \APACyear {2023}}%
}]{%
zero-shot}
\APACinsertmetastar {%
zero-shot}%
\begin{APACrefauthors}%
Gilardi, F.%
, Alizadeh, M.%
\BCBL {}\ \BBA {} Kubli, M.%
\end{APACrefauthors}%
\unskip\
\newblock
\APACrefYearMonthDay{2023}{}{}.
\newblock
{\BBOQ}\APACrefatitle {ChatGPT Outperforms Crowd-Workers for Text-Annotation Tasks} {Chatgpt outperforms crowd-workers for text-annotation tasks}.{\BBCQ}
\newblock
\APACjournalVolNumPages{Proceedings of the National Academy of Sciences}{}{}{}.
\PrintBackRefs{\CurrentBib}

\bibitem [\protect \citeauthoryear {%
Goodfellow%
, Bengio%
\BCBL {}\ \BBA {} Courville%
}{%
Goodfellow%
\ \protect \BOthers {.}}{%
{\protect \APACyear {2016}}%
}]{%
deep_learning}
\APACinsertmetastar {%
deep_learning}%
\begin{APACrefauthors}%
Goodfellow, I.%
, Bengio, Y.%
\BCBL {}\ \BBA {} Courville, A.%
\end{APACrefauthors}%
\unskip\
\newblock
\APACrefYear{2016}.
\newblock
\APACrefbtitle {Deep Learning} {Deep learning}.
\newblock
\APACaddressPublisher{}{The MIT Press}.
\PrintBackRefs{\CurrentBib}

\bibitem [\protect \citeauthoryear {%
Hastie%
, Tibshirani%
\BCBL {}\ \BBA {} Friedman%
}{%
Hastie%
\ \protect \BOthers {.}}{%
{\protect \APACyear {2009}}%
}]{%
elements}
\APACinsertmetastar {%
elements}%
\begin{APACrefauthors}%
Hastie, T.%
, Tibshirani, R.%
\BCBL {}\ \BBA {} Friedman, J.%
\end{APACrefauthors}%
\unskip\
\newblock
\APACrefYear{2009}.
\newblock
\APACrefbtitle {The Elements of Statistical Learning: Data Mining, Inference, and Prediction} {The elements of statistical learning: Data mining, inference, and prediction}.
\newblock
\APACaddressPublisher{}{Springer}.
\PrintBackRefs{\CurrentBib}

\bibitem [\protect \citeauthoryear {%
Hu%
\ \protect \BOthers {.}}{%
Hu%
\ \protect \BOthers {.}}{%
{\protect \APACyear {2022}}%
}]{%
conflibert}
\APACinsertmetastar {%
conflibert}%
\begin{APACrefauthors}%
Hu, Y.%
, Hosseini, M.%
, Parolin, E\BPBI S.%
, Osorio, J.%
, Khan, L.%
, Brandt, P\BPBI T.%
\BCBL {}\ \BBA {} D'Orazio, V\BPBI J.%
\end{APACrefauthors}%
\unskip\
\newblock
\APACrefYearMonthDay{2022}{}{}.
\newblock
{\BBOQ}\APACrefatitle {ConfliBERT: A Pre-trained Language Model for Political Conflict and Violence} {Conflibert: A pre-trained language model for political conflict and violence}.{\BBCQ}
\newblock
\APACjournalVolNumPages{Proceedings of the 2022 Conference of the North American Chapter of the Association for Computational Linguistics}{}{}{}.
\PrintBackRefs{\CurrentBib}

\bibitem [\protect \citeauthoryear {%
Huang%
, Kwak%
\BCBL {}\ \BBA {} An%
}{%
Huang%
\ \protect \BOthers {.}}{%
{\protect \APACyear {2023}}%
}]{%
gpt-hate}
\APACinsertmetastar {%
gpt-hate}%
\begin{APACrefauthors}%
Huang, F.%
, Kwak, H.%
\BCBL {}\ \BBA {} An, J.%
\end{APACrefauthors}%
\unskip\
\newblock
\APACrefYearMonthDay{2023}{}{}.
\newblock
{\BBOQ}\APACrefatitle {Is ChatGPT better than Human Annotators? Potential and Limitations of ChatGPT in Explaining Implicit Hate Speech} {Is chatgpt better than human annotators? potential and limitations of chatgpt in explaining implicit hate speech}.{\BBCQ}
\newblock
\APACjournalVolNumPages{WWW '23 Companion: Companion Proceedings of the ACM Web Conference 2023}{}{}{}.
\PrintBackRefs{\CurrentBib}

\bibitem [\protect \citeauthoryear {%
Häffner%
, Hofer%
, Nagl%
\BCBL {}\ \BBA {} Walterskirchen%
}{%
Häffner%
\ \protect \BOthers {.}}{%
{\protect \APACyear {2023}}%
}]{%
interpretable}
\APACinsertmetastar {%
interpretable}%
\begin{APACrefauthors}%
Häffner, S.%
, Hofer, M.%
, Nagl, M.%
\BCBL {}\ \BBA {} Walterskirchen, J.%
\end{APACrefauthors}%
\unskip\
\newblock
\APACrefYearMonthDay{2023}{}{}.
\newblock
{\BBOQ}\APACrefatitle {Introducing an Interpretable Deep Learning Approach to Domain-Specific Dictionary Creation: A Use Case for Conflict Prediction} {Introducing an interpretable deep learning approach to domain-specific dictionary creation: A use case for conflict prediction}.{\BBCQ}
\newblock
\APACjournalVolNumPages{Political Analysis}{}{}{1–19}.
\newblock
\begin{APACrefDOI} \doi{10.1017/pan.2023.7} \end{APACrefDOI}
\PrintBackRefs{\CurrentBib}

\bibitem [\protect \citeauthoryear {%
Kaufman%
}{%
Kaufman%
}{%
{\protect \APACyear {2024}}%
}]{%
sample_selection}
\APACinsertmetastar {%
sample_selection}%
\begin{APACrefauthors}%
Kaufman, A\BPBI R.%
\end{APACrefauthors}%
\unskip\
\newblock
\APACrefYearMonthDay{2024}{}{}.
\newblock
{\BBOQ}\APACrefatitle {Selecting More Informative Training Sets with Fewer Observations} {Selecting more informative training sets with fewer observations}.{\BBCQ}
\newblock
\APACjournalVolNumPages{Political Analysis}{}{}{}.
\PrintBackRefs{\CurrentBib}

\bibitem [\protect \citeauthoryear {%
Kaufman%
\ \BBA {} Klevs%
}{%
Kaufman%
\ \BBA {} Klevs%
}{%
{\protect \APACyear {2022}}%
}]{%
fuzzy}
\APACinsertmetastar {%
fuzzy}%
\begin{APACrefauthors}%
Kaufman, A\BPBI R.%
\BCBT {}\ \BBA {} Klevs, A.%
\end{APACrefauthors}%
\unskip\
\newblock
\APACrefYearMonthDay{2022}{}{}.
\newblock
{\BBOQ}\APACrefatitle {Adaptive Fuzzy String Matching: How to Merge Datasets with Only One (Messy) Identifying Field} {Adaptive fuzzy string matching: How to merge datasets with only one (messy) identifying field}.{\BBCQ}
\newblock
\APACjournalVolNumPages{Political Analysis}{}{}{}.
\PrintBackRefs{\CurrentBib}

\bibitem [\protect \citeauthoryear {%
Kim%
}{%
Kim%
}{%
{\protect \APACyear {2022}}%
}]{%
rhetoric_on_twitter}
\APACinsertmetastar {%
rhetoric_on_twitter}%
\begin{APACrefauthors}%
Kim, T.%
\end{APACrefauthors}%
\unskip\
\newblock
\APACrefYearMonthDay{2022}{}{}.
\newblock
{\BBOQ}\APACrefatitle {Violent Political rhetoric on Twitter} {Violent political rhetoric on twitter}.{\BBCQ}
\newblock
\APACjournalVolNumPages{Political Science Research and Methods}{}{}{}.
\PrintBackRefs{\CurrentBib}

\bibitem [\protect \citeauthoryear {%
O.~Kjell%
, Giorgi%
\BCBL {}\ \BBA {} Schwartz%
}{%
O.~Kjell%
\ \protect \BOthers {.}}{%
{\protect \APACyear {2023}}%
}]{%
text_package}
\APACinsertmetastar {%
text_package}%
\begin{APACrefauthors}%
Kjell, O.%
, Giorgi, S.%
\BCBL {}\ \BBA {} Schwartz, H\BPBI A.%
\end{APACrefauthors}%
\unskip\
\newblock
\APACrefYearMonthDay{2023}{}{}.
\newblock
{\BBOQ}\APACrefatitle {The Text-Package: An R-Package for Analyzing and Visualizing Human Language Using Normal Language Processing and Transformers} {The text-package: An r-package for analyzing and visualizing human language using normal language processing and transformers}.{\BBCQ}
\newblock
\APACjournalVolNumPages{Psychological Methods}{}{}{}.
\PrintBackRefs{\CurrentBib}

\bibitem [\protect \citeauthoryear {%
O\BPBI N.~Kjell%
, Kjell%
\BCBL {}\ \BBA {} Schwartz%
}{%
O\BPBI N.~Kjell%
\ \protect \BOthers {.}}{%
{\protect \APACyear {2024}}%
}]{%
beyond_rating_scales}
\APACinsertmetastar {%
beyond_rating_scales}%
\begin{APACrefauthors}%
Kjell, O\BPBI N.%
, Kjell, K.%
\BCBL {}\ \BBA {} Schwartz, H\BPBI A.%
\end{APACrefauthors}%
\unskip\
\newblock
\APACrefYearMonthDay{2024}{}{}.
\newblock
{\BBOQ}\APACrefatitle {Beyond Rating Scales: With Targeted Evaluation, Large Language Models Are Poised for Psychological Assessment} {Beyond rating scales: With targeted evaluation, large language models are poised for psychological assessment}.{\BBCQ}
\newblock
\APACjournalVolNumPages{Psychiatry Research}{}{}{}.
\PrintBackRefs{\CurrentBib}

\bibitem [\protect \citeauthoryear {%
Korinek%
}{%
Korinek%
}{%
{\protect \APACyear {2023}}%
}]{%
genai_for_economists}
\APACinsertmetastar {%
genai_for_economists}%
\begin{APACrefauthors}%
Korinek, A.%
\end{APACrefauthors}%
\unskip\
\newblock
\APACrefYearMonthDay{2023}{}{}.
\newblock
{\BBOQ}\APACrefatitle {Generative AI for Economic Research: Use Cases and Implications for Economists} {Generative ai for economic research: Use cases and implications for economists}.{\BBCQ}
\newblock
\APACjournalVolNumPages{Journal of Economic Literature}{}{}{}.
\PrintBackRefs{\CurrentBib}

\bibitem [\protect \citeauthoryear {%
Lai%
\ \protect \BOthers {.}}{%
Lai%
\ \protect \BOthers {.}}{%
{\protect \APACyear {2024}}%
}]{%
youtube_ideology}
\APACinsertmetastar {%
youtube_ideology}%
\begin{APACrefauthors}%
Lai, A.%
, Brown, M\BPBI A.%
, Bisbee, J.%
, Tucker, J\BPBI A.%
, Nagler, J.%
\BCBL {}\ \BBA {} Bonneau, R.%
\end{APACrefauthors}%
\unskip\
\newblock
\APACrefYearMonthDay{2024}{}{}.
\newblock
{\BBOQ}\APACrefatitle {Estimating the Ideology of Political YouTube Videos} {Estimating the ideology of political youtube videos}.{\BBCQ}
\newblock
\APACjournalVolNumPages{Political Analysis}{}{}{}.
\PrintBackRefs{\CurrentBib}

\bibitem [\protect \citeauthoryear {%
Lan%
\ \protect \BOthers {.}}{%
Lan%
\ \protect \BOthers {.}}{%
{\protect \APACyear {2020}}%
}]{%
albert}
\APACinsertmetastar {%
albert}%
\begin{APACrefauthors}%
Lan, Z.%
, Chen, M.%
, Goodman, S.%
, Gimpel, K.%
, Sharma, P.%
\BCBL {}\ \BBA {} Soricut, R.%
\end{APACrefauthors}%
\unskip\
\newblock
\APACrefYearMonthDay{2020}{}{}.
\newblock
{\BBOQ}\APACrefatitle {{ALBERT: A Lite BERT for Self-supervised Learning of Language Representations}} {{ALBERT: A Lite BERT for Self-supervised Learning of Language Representations}}.{\BBCQ}
\newblock
\APACjournalVolNumPages{ICLR}{}{}{}.
\PrintBackRefs{\CurrentBib}

\bibitem [\protect \citeauthoryear {%
Laurer%
, van Atteveldt%
, Casas%
\BCBL {}\ \BBA {} Welbers%
}{%
Laurer%
\ \protect \BOthers {.}}{%
{\protect \APACyear {2024}}%
}]{%
bert_nli}
\APACinsertmetastar {%
bert_nli}%
\begin{APACrefauthors}%
Laurer, M.%
, van Atteveldt, W.%
, Casas, A.%
\BCBL {}\ \BBA {} Welbers, K.%
\end{APACrefauthors}%
\unskip\
\newblock
\APACrefYearMonthDay{2024}{}{}.
\newblock
{\BBOQ}\APACrefatitle {Less Annotating, More Classifying: Addressing the Data Scarcity Issue of Supervised Machine Learning with Deep Transfer Learning and BERT-NLI} {Less annotating, more classifying: Addressing the data scarcity issue of supervised machine learning with deep transfer learning and bert-nli}.{\BBCQ}
\newblock
\APACjournalVolNumPages{Political Analysis}{}{}{}.
\PrintBackRefs{\CurrentBib}

\bibitem [\protect \citeauthoryear {%
Lee%
\ \protect \BOthers {.}}{%
Lee%
\ \protect \BOthers {.}}{%
{\protect \APACyear {2019}}%
}]{%
biobert}
\APACinsertmetastar {%
biobert}%
\begin{APACrefauthors}%
Lee, J.%
, Yoon, W.%
, Kim, S.%
, Kim, D.%
, Kim, S.%
, So, C\BPBI H.%
\BCBL {}\ \BBA {} Kang, J.%
\end{APACrefauthors}%
\unskip\
\newblock
\APACrefYearMonthDay{2019}{}{}.
\newblock
{\BBOQ}\APACrefatitle {Biobert: A Pre-Trained Biomedical Language Representation Model For Biomedical Text Mining} {Biobert: A pre-trained biomedical language representation model for biomedical text mining}.{\BBCQ}
\newblock
\APACjournalVolNumPages{Bioinformatics}{36}{}{}.
\PrintBackRefs{\CurrentBib}

\bibitem [\protect \citeauthoryear {%
J.~Liu%
\ \protect \BOthers {.}}{%
J.~Liu%
\ \protect \BOthers {.}}{%
{\protect \APACyear {2022}}%
}]{%
liu2022what}
\APACinsertmetastar {%
liu2022what}%
\begin{APACrefauthors}%
Liu, J.%
, Shen, D.%
, Zhang, Y.%
, Dolan, B.%
, Carin, L.%
\BCBL {}\ \BBA {} Chen, W.%
\end{APACrefauthors}%
\unskip\
\newblock
\APACrefYearMonthDay{2022}{January}{}.
\newblock
{\BBOQ}\APACrefatitle {What Makes Good In-Context Examples for GPT-3?} {What makes good in-context examples for gpt-3?}{\BBCQ}
\newblock
\BIn{} \APACrefbtitle {DeeLIO 2022 - Deep Learning Inside Out: 3rd Workshop on Knowledge Extraction and Integration for Deep Learning Architectures, Proceedings of the Workshop.} {Deelio 2022 - deep learning inside out: 3rd workshop on knowledge extraction and integration for deep learning architectures, proceedings of the workshop.}
\PrintBackRefs{\CurrentBib}

\bibitem [\protect \citeauthoryear {%
Y.~Liu%
\ \protect \BOthers {.}}{%
Y.~Liu%
\ \protect \BOthers {.}}{%
{\protect \APACyear {2019}}%
}]{%
roberta}
\APACinsertmetastar {%
roberta}%
\begin{APACrefauthors}%
Liu, Y.%
, Ott, M.%
, Goyal, N.%
, Du, J.%
, Joshi, M.%
, Chen, D.%
\BDBL {}Stoyanov, V.%
\end{APACrefauthors}%
\unskip\
\newblock
\APACrefYearMonthDay{2019}{}{}.
\newblock
{\BBOQ}\APACrefatitle {{RoBERTa: A Robustly Optimized BERT Pretraining Approach}} {{RoBERTa: A Robustly Optimized BERT Pretraining Approach}}.{\BBCQ}
\newblock
\APACjournalVolNumPages{arXiv:1907.11692}{}{}{}.
\PrintBackRefs{\CurrentBib}

\bibitem [\protect \citeauthoryear {%
Longpre%
, Wang%
\BCBL {}\ \BBA {} DuBois%
}{%
Longpre%
\ \protect \BOthers {.}}{%
{\protect \APACyear {2020}}%
}]{%
aug}
\APACinsertmetastar {%
aug}%
\begin{APACrefauthors}%
Longpre, S.%
, Wang, Y.%
\BCBL {}\ \BBA {} DuBois, C.%
\end{APACrefauthors}%
\unskip\
\newblock
\APACrefYearMonthDay{2020}{}{}.
\newblock
{\BBOQ}\APACrefatitle {{How Effective is Task-Agnostic Data Augmentation for Pretrained Transformers?}} {{How Effective is Task-Agnostic Data Augmentation for Pretrained Transformers?}}{\BBCQ}
\newblock
\APACjournalVolNumPages{Findings of the Association for Computational Linguistics: EMNLP 2020}{}{}{}.
\PrintBackRefs{\CurrentBib}

\bibitem [\protect \citeauthoryear {%
Mei%
, Xie%
, Yuan%
\BCBL {}\ \BBA {} Jackson%
}{%
Mei%
\ \protect \BOthers {.}}{%
{\protect \APACyear {2024}}%
}]{%
turing_test_ai}
\APACinsertmetastar {%
turing_test_ai}%
\begin{APACrefauthors}%
Mei, Q.%
, Xie, Y.%
, Yuan, W.%
\BCBL {}\ \BBA {} Jackson, M\BPBI O.%
\end{APACrefauthors}%
\unskip\
\newblock
\APACrefYearMonthDay{2024}{}{}.
\newblock
{\BBOQ}\APACrefatitle {A Turing Test of Whether AI Chatbots are Behaviorally Similar to Humans} {A turing test of whether ai chatbots are behaviorally similar to humans}.{\BBCQ}
\newblock
\APACjournalVolNumPages{Proceedings of the National Academy of Sciences}{}{}{}.
\PrintBackRefs{\CurrentBib}

\bibitem [\protect \citeauthoryear {%
Mikolov%
, Sutskever%
, Chen%
, Corrado%
\BCBL {}\ \BBA {} Dean%
}{%
Mikolov%
\ \protect \BOthers {.}}{%
{\protect \APACyear {2013}}%
}]{%
word2vec}
\APACinsertmetastar {%
word2vec}%
\begin{APACrefauthors}%
Mikolov, T.%
, Sutskever, I.%
, Chen, K.%
, Corrado, G.%
\BCBL {}\ \BBA {} Dean, J.%
\end{APACrefauthors}%
\unskip\
\newblock
\APACrefYearMonthDay{2013}{}{}.
\newblock
{\BBOQ}\APACrefatitle {Distributed Representations of Words and Phrases and their Compositionality} {Distributed representations of words and phrases and their compositionality}.{\BBCQ}
\newblock
\APACjournalVolNumPages{NIPS'13: Proceedings of the 26th International Conference on Neural Information Processing Systems}{}{}{}.
\PrintBackRefs{\CurrentBib}

\bibitem [\protect \citeauthoryear {%
Muchlinski%
, Siroky%
, He%
\BCBL {}\ \BBA {} Kocher%
}{%
Muchlinski%
\ \protect \BOthers {.}}{%
{\protect \APACyear {2016}}%
}]{%
predicted_acc1}
\APACinsertmetastar {%
predicted_acc1}%
\begin{APACrefauthors}%
Muchlinski, D.%
, Siroky, D.%
, He, J.%
\BCBL {}\ \BBA {} Kocher, M.%
\end{APACrefauthors}%
\unskip\
\newblock
\APACrefYearMonthDay{2016}{}{}.
\newblock
{\BBOQ}\APACrefatitle {{Comparing Random Forest with Logistic Regression for Predicting Class-Imbalanced Civil War Onset Data}} {{Comparing Random Forest with Logistic Regression for Predicting Class-Imbalanced Civil War Onset Data}}.{\BBCQ}
\newblock
\APACjournalVolNumPages{Political Analysis}{24}{1}{87-103}.
\PrintBackRefs{\CurrentBib}

\bibitem [\protect \citeauthoryear {%
Nielbo%
\ \protect \BOthers {.}}{%
Nielbo%
\ \protect \BOthers {.}}{%
{\protect \APACyear {2024}}%
}]{%
qta}
\APACinsertmetastar {%
qta}%
\begin{APACrefauthors}%
Nielbo, K\BPBI L.%
, Karsdorp, F.%
, Wevers, M.%
, Lassche, A.%
, Baglini, R\BPBI B.%
, Kestemont, M.%
\BCBL {}\ \BBA {} Tahmasebi, N.%
\end{APACrefauthors}%
\unskip\
\newblock
\APACrefYearMonthDay{2024}{April}{}.
\newblock
{\BBOQ}\APACrefatitle {Quantitative Text Analysis} {Quantitative text analysis}.{\BBCQ}
\newblock
\APACjournalVolNumPages{Nature Reviews Methods Primers}{4}{}{Article 25}.
\PrintBackRefs{\CurrentBib}

\bibitem [\protect \citeauthoryear {%
Osnabrügge%
, Ash%
\BCBL {}\ \BBA {} Morelli%
}{%
Osnabrügge%
\ \protect \BOthers {.}}{%
{\protect \APACyear {2021}}%
}]{%
cross_domain}
\APACinsertmetastar {%
cross_domain}%
\begin{APACrefauthors}%
Osnabrügge, M.%
, Ash, E.%
\BCBL {}\ \BBA {} Morelli, M.%
\end{APACrefauthors}%
\unskip\
\newblock
\APACrefYearMonthDay{2021}{}{}.
\newblock
{\BBOQ}\APACrefatitle {Cross-Domain Topic Classification for Political Texts} {Cross-domain topic classification for political texts}.{\BBCQ}
\newblock
\APACjournalVolNumPages{Political Analysis}{}{}{}.
\PrintBackRefs{\CurrentBib}

\bibitem [\protect \citeauthoryear {%
Peng%
, Qiu%
, Fosse%
\BCBL {}\ \BBA {} Uzzi%
}{%
Peng%
\ \protect \BOthers {.}}{%
{\protect \APACyear {2024}}%
}]{%
promotional_language}
\APACinsertmetastar {%
promotional_language}%
\begin{APACrefauthors}%
Peng, H.%
, Qiu, H\BPBI S.%
, Fosse, H\BPBI B.%
\BCBL {}\ \BBA {} Uzzi, B.%
\end{APACrefauthors}%
\unskip\
\newblock
\APACrefYearMonthDay{2024}{}{}.
\newblock
{\BBOQ}\APACrefatitle {Promotional Language and the Adoption of Innovative Ideas in Science} {Promotional language and the adoption of innovative ideas in science}.{\BBCQ}
\newblock
\APACjournalVolNumPages{Proceedings of the National Academy of Sciences of the United States of America}{}{}{}.
\PrintBackRefs{\CurrentBib}

\bibitem [\protect \citeauthoryear {%
Pennington%
, Socher%
\BCBL {}\ \BBA {} Manning%
}{%
Pennington%
\ \protect \BOthers {.}}{%
{\protect \APACyear {2014}}%
}]{%
glove}
\APACinsertmetastar {%
glove}%
\begin{APACrefauthors}%
Pennington, J.%
, Socher, R.%
\BCBL {}\ \BBA {} Manning, C.%
\end{APACrefauthors}%
\unskip\
\newblock
\APACrefYearMonthDay{2014}{}{}.
\newblock
{\BBOQ}\APACrefatitle {GloVe: Global Vectors for Word Representation} {Glove: Global vectors for word representation}.{\BBCQ}
\newblock
\APACjournalVolNumPages{Proceedings of the 2014 Conference on Empirical Methods in Natural Language Processing (EMNLP)}{}{}{}.
\PrintBackRefs{\CurrentBib}

\bibitem [\protect \citeauthoryear {%
Radford%
\ \protect \BOthers {.}}{%
Radford%
\ \protect \BOthers {.}}{%
{\protect \APACyear {2019}}%
}]{%
radford2019language}
\APACinsertmetastar {%
radford2019language}%
\begin{APACrefauthors}%
Radford, A.%
, Wu, J.%
, Child, R.%
, Luan, D.%
, Amodei, D.%
\BCBL {}\ \BBA {} Sutskever, I.%
\end{APACrefauthors}%
\unskip\
\newblock
\APACrefYearMonthDay{2019}{}{}.
\newblock
{\BBOQ}\APACrefatitle {Language Models are Unsupervised Multitask Learners} {Language models are unsupervised multitask learners}.{\BBCQ}
\newblock
\APACjournalVolNumPages{OpenAI}{}{}{}.
\newblock
\begin{APACrefURL} \url{https://www.openai.com/research/language-models-are-unsupervised-multitask-learners} \end{APACrefURL}
\PrintBackRefs{\CurrentBib}

\bibitem [\protect \citeauthoryear {%
Rai%
\ \protect \BOthers {.}}{%
Rai%
\ \protect \BOthers {.}}{%
{\protect \APACyear {2024}}%
}]{%
language_marker}
\APACinsertmetastar {%
language_marker}%
\begin{APACrefauthors}%
Rai, S.%
, Stade, E\BPBI C.%
, Giorgi, S.%
, Francisco, A.%
, Ungar, L\BPBI H.%
, Curtis, B.%
\BCBL {}\ \BBA {} Guntuku, S\BPBI C.%
\end{APACrefauthors}%
\unskip\
\newblock
\APACrefYearMonthDay{2024}{}{}.
\newblock
{\BBOQ}\APACrefatitle {Key language markers of depression on social media depend on race} {Key language markers of depression on social media depend on race}.{\BBCQ}
\newblock
\APACjournalVolNumPages{Proceedings of the National Academy of Sciences}{121}{14}{e2319837121}.
\newblock
\begin{APACrefURL} \url{https://www.pnas.org/doi/abs/10.1073/pnas.2319837121} \end{APACrefURL}
\newblock
\begin{APACrefDOI} \doi{10.1073/pnas.2319837121} \end{APACrefDOI}
\PrintBackRefs{\CurrentBib}

\bibitem [\protect \citeauthoryear {%
Rheault%
\ \BBA {} Cochrane%
}{%
Rheault%
\ \BBA {} Cochrane%
}{%
{\protect \APACyear {2019}}%
}]{%
ideological_placement}
\APACinsertmetastar {%
ideological_placement}%
\begin{APACrefauthors}%
Rheault, L.%
\BCBT {}\ \BBA {} Cochrane, C.%
\end{APACrefauthors}%
\unskip\
\newblock
\APACrefYearMonthDay{2019}{}{}.
\newblock
{\BBOQ}\APACrefatitle {Word Embeddings for the Analysis of Ideological Placement in Parliamentary Corpora} {Word embeddings for the analysis of ideological placement in parliamentary corpora}.{\BBCQ}
\newblock
\APACjournalVolNumPages{Political Analysis}{}{}{}.
\PrintBackRefs{\CurrentBib}

\bibitem [\protect \citeauthoryear {%
Rodman%
}{%
Rodman%
}{%
{\protect \APACyear {2019}}%
}]{%
political_concepts}
\APACinsertmetastar {%
political_concepts}%
\begin{APACrefauthors}%
Rodman, E.%
\end{APACrefauthors}%
\unskip\
\newblock
\APACrefYearMonthDay{2019}{}{}.
\newblock
{\BBOQ}\APACrefatitle {A Timely Intervention: Tracking the Changing Meanings of Political Concepts with Word Vectors} {A timely intervention: Tracking the changing meanings of political concepts with word vectors}.{\BBCQ}
\newblock
\APACjournalVolNumPages{Political Analysis}{}{}{}.
\PrintBackRefs{\CurrentBib}

\bibitem [\protect \citeauthoryear {%
Rodriguez%
\ \BBA {} Spirling%
}{%
Rodriguez%
\ \BBA {} Spirling%
}{%
{\protect \APACyear {2022}}%
}]{%
embedding_discussion}
\APACinsertmetastar {%
embedding_discussion}%
\begin{APACrefauthors}%
Rodriguez, P\BPBI L.%
\BCBT {}\ \BBA {} Spirling, A.%
\end{APACrefauthors}%
\unskip\
\newblock
\APACrefYearMonthDay{2022}{}{}.
\newblock
{\BBOQ}\APACrefatitle {Word Embeddings What Works, What Doesn'T, and How To Tell the Difference For Applied Research} {Word embeddings what works, what doesn't, and how to tell the difference for applied research}.{\BBCQ}
\newblock
\APACjournalVolNumPages{Journal of Politics}{}{}{}.
\PrintBackRefs{\CurrentBib}

\bibitem [\protect \citeauthoryear {%
Simchon%
, Hadar%
\BCBL {}\ \BBA {} Gilead%
}{%
Simchon%
\ \protect \BOthers {.}}{%
{\protect \APACyear {2023}}%
}]{%
linguistic_agency}
\APACinsertmetastar {%
linguistic_agency}%
\begin{APACrefauthors}%
Simchon, A.%
, Hadar, B.%
\BCBL {}\ \BBA {} Gilead, M.%
\end{APACrefauthors}%
\unskip\
\newblock
\APACrefYearMonthDay{2023}{}{}.
\newblock
{\BBOQ}\APACrefatitle {A Computational Text Analysis Investigation of the Relation between Personal and Linguistic Agency} {A computational text analysis investigation of the relation between personal and linguistic agency}.{\BBCQ}
\newblock
\APACjournalVolNumPages{Communications Psychology}{}{}{}.
\PrintBackRefs{\CurrentBib}

\bibitem [\protect \citeauthoryear {%
Strachan%
\ \protect \BOthers {.}}{%
Strachan%
\ \protect \BOthers {.}}{%
{\protect \APACyear {2024}}%
}]{%
theory_of_mind}
\APACinsertmetastar {%
theory_of_mind}%
\begin{APACrefauthors}%
Strachan, J\BPBI W\BPBI A.%
, Albergo, D.%
, Borghini, G.%
, Pansardi, O.%
, Scaliti, E.%
, Gupta, S.%
\BDBL {}Becchio, C.%
\end{APACrefauthors}%
\unskip\
\newblock
\APACrefYearMonthDay{2024}{}{}.
\newblock
{\BBOQ}\APACrefatitle {Testing Theory of Mind in Large Language Models and Humans} {Testing theory of mind in large language models and humans}.{\BBCQ}
\newblock
\APACjournalVolNumPages{Nature Human Behaviour}{}{}{}.
\PrintBackRefs{\CurrentBib}

\bibitem [\protect \citeauthoryear {%
Torres%
}{%
Torres%
}{%
{\protect \APACyear {2023}}%
}]{%
unsupervised_semi_supervised_visual_frames}
\APACinsertmetastar {%
unsupervised_semi_supervised_visual_frames}%
\begin{APACrefauthors}%
Torres, M.%
\end{APACrefauthors}%
\unskip\
\newblock
\APACrefYearMonthDay{2023}{}{}.
\newblock
{\BBOQ}\APACrefatitle {A Framework for the Unsupervised and Semi-Supervised Analysis of Visual Frames} {A framework for the unsupervised and semi-supervised analysis of visual frames}.{\BBCQ}
\newblock
\APACjournalVolNumPages{Political Analysis}{}{}{}.
\PrintBackRefs{\CurrentBib}

\bibitem [\protect \citeauthoryear {%
Vaswani%
\ \protect \BOthers {.}}{%
Vaswani%
\ \protect \BOthers {.}}{%
{\protect \APACyear {2017}}%
}]{%
attention}
\APACinsertmetastar {%
attention}%
\begin{APACrefauthors}%
Vaswani, A.%
, Shazeer, N.%
, Parmar, N.%
, Uszkoreit, J.%
, Jones, L.%
, Gomez, A\BPBI N.%
\BDBL {}Polosukhin, I.%
\end{APACrefauthors}%
\unskip\
\newblock
\APACrefYearMonthDay{2017}{}{}.
\newblock
{\BBOQ}\APACrefatitle {{Attention Is All You Need}} {{Attention Is All You Need}}.{\BBCQ}
\newblock
\APACjournalVolNumPages{31st Conference on Neural Information Processing Systems}{}{}{}.
\PrintBackRefs{\CurrentBib}

\bibitem [\protect \citeauthoryear {%
X.~Wang%
\ \protect \BOthers {.}}{%
X.~Wang%
\ \protect \BOthers {.}}{%
{\protect \APACyear {2023}}%
}]{%
wang2023self}
\APACinsertmetastar {%
wang2023self}%
\begin{APACrefauthors}%
Wang, X.%
, Wei, J.%
, Schuurmans, D.%
, Le, Q.%
, Chi, E\BPBI H.%
, Narang, S.%
\BDBL {}Zhou, D.%
\end{APACrefauthors}%
\unskip\
\newblock
\APACrefYearMonthDay{2023}{}{}.
\newblock
{\BBOQ}\APACrefatitle {Self-Consistency Improves Chain of Thought Reasoning in Language Models} {Self-consistency improves chain of thought reasoning in language models}.{\BBCQ}
\newblock
\BIn{} \APACrefbtitle {{Proceedings of the International Conference on Learning Representations (ICLR 2023)}.} {{Proceedings of the International Conference on Learning Representations (ICLR 2023)}.}
\PrintBackRefs{\CurrentBib}

\bibitem [\protect \citeauthoryear {%
Y.~Wang%
}{%
Y.~Wang%
}{%
{\protect \APACyear {2019}}%
{\protect \APACexlab {{\protect \BCnt {1}}}}}]{%
predicted_acc2}
\APACinsertmetastar {%
predicted_acc2}%
\begin{APACrefauthors}%
Wang, Y.%
\end{APACrefauthors}%
\unskip\
\newblock
\APACrefYearMonthDay{2019{\protect \BCnt {1}}}{}{}.
\newblock
{\BBOQ}\APACrefatitle {{Comparing Random Forest with Logistic Regression for Predicting Class-Imbalanced Civil War Onset Data: A Comment}} {{Comparing Random Forest with Logistic Regression for Predicting Class-Imbalanced Civil War Onset Data: A Comment}}.{\BBCQ}
\newblock
\APACjournalVolNumPages{Political Analysis}{21}{1}{107-110}.
\PrintBackRefs{\CurrentBib}

\bibitem [\protect \citeauthoryear {%
Y.~Wang%
}{%
Y.~Wang%
}{%
{\protect \APACyear {2019}}%
{\protect \APACexlab {{\protect \BCnt {2}}}}}]{%
pca}
\APACinsertmetastar {%
pca}%
\begin{APACrefauthors}%
Wang, Y.%
\end{APACrefauthors}%
\unskip\
\newblock
\APACrefYearMonthDay{2019{\protect \BCnt {2}}}{}{}.
\newblock
{\BBOQ}\APACrefatitle {Single Training Dimension Selection for Word Embedding with PCA} {Single training dimension selection for word embedding with pca}.{\BBCQ}
\newblock
\APACjournalVolNumPages{Proceedings of the 2019 Conference on Empirical Methods in Natural Language Processing and the 9th International Joint Conference on Natural Language Processing (EMNLP-IJCNLP)}{}{}{}.
\PrintBackRefs{\CurrentBib}

\bibitem [\protect \citeauthoryear {%
Y.~Wang%
}{%
Y.~Wang%
}{%
{\protect \APACyear {2023}}%
{\protect \APACexlab {{\protect \BCnt {1}}}}}]{%
finetune_pa}
\APACinsertmetastar {%
finetune_pa}%
\begin{APACrefauthors}%
Wang, Y.%
\end{APACrefauthors}%
\unskip\
\newblock
\APACrefYearMonthDay{2023{\protect \BCnt {1}}}{}{}.
\newblock
{\BBOQ}\APACrefatitle {On Finetuning Large Language Models} {On finetuning large language models}.{\BBCQ}
\newblock
\APACjournalVolNumPages{Political Analysis}{}{}{}.
\PrintBackRefs{\CurrentBib}

\bibitem [\protect \citeauthoryear {%
Y.~Wang%
}{%
Y.~Wang%
}{%
{\protect \APACyear {2023}}%
{\protect \APACexlab {{\protect \BCnt {2}}}}}]{%
pretrained_topic_classification}
\APACinsertmetastar {%
pretrained_topic_classification}%
\begin{APACrefauthors}%
Wang, Y.%
\end{APACrefauthors}%
\unskip\
\newblock
\APACrefYearMonthDay{2023{\protect \BCnt {2}}}{}{}.
\newblock
{\BBOQ}\APACrefatitle {Topic Classification for Political Texts with Pretrained Language Models} {Topic classification for political texts with pretrained language models}.{\BBCQ}
\newblock
\APACjournalVolNumPages{Political Analysis}{}{}{}.
\PrintBackRefs{\CurrentBib}

\bibitem [\protect \citeauthoryear {%
Y.~Wang%
}{%
Y.~Wang%
}{%
{\protect \APACyear {2024}}%
}]{%
depression_prediction}
\APACinsertmetastar {%
depression_prediction}%
\begin{APACrefauthors}%
Wang, Y.%
\end{APACrefauthors}%
\unskip\
\newblock
\APACrefYearMonthDay{2024}{}{}.
\newblock
{\BBOQ}\APACrefatitle {Large Language Models for Depression Prediction} {Large language models for depression prediction}.{\BBCQ}
\newblock
\APACjournalVolNumPages{Proceedings of the National Academy of Sciences}{}{}{}.
\PrintBackRefs{\CurrentBib}

\bibitem [\protect \citeauthoryear {%
Y.~Wang%
, Feng%
, Hong%
, Berger%
\BCBL {}\ \BBA {} Luo%
}{%
Y.~Wang%
\ \protect \BOthers {.}}{%
{\protect \APACyear {2017}}%
}]{%
polarization_trump_clinton_followers}
\APACinsertmetastar {%
polarization_trump_clinton_followers}%
\begin{APACrefauthors}%
Wang, Y.%
, Feng, Y.%
, Hong, Z.%
, Berger, R.%
\BCBL {}\ \BBA {} Luo, J.%
\end{APACrefauthors}%
\unskip\
\newblock
\APACrefYearMonthDay{2017}{}{}.
\newblock
{\BBOQ}\APACrefatitle {How Polarized Have We Become? A Multimodal Classification of Trump Followers and Clinton Followers} {How polarized have we become? a multimodal classification of trump followers and clinton followers}.{\BBCQ}
\newblock
\APACjournalVolNumPages{Social Informatics}{}{}{}.
\PrintBackRefs{\CurrentBib}

\bibitem [\protect \citeauthoryear {%
Y.~Wang%
\ \BBA {} Qu%
}{%
Y.~Wang%
\ \BBA {} Qu%
}{%
{\protect \APACyear {2024}}%
}]{%
tutorial}
\APACinsertmetastar {%
tutorial}%
\begin{APACrefauthors}%
Wang, Y.%
\BCBT {}\ \BBA {} Qu, W.%
\end{APACrefauthors}%
\unskip\
\newblock
\APACrefYearMonthDay{2024}{}{}.
\newblock
{\BBOQ}\APACrefatitle {A Tutorial on the Pretrain-Finetune Paradigm for Natural Language Processing} {A tutorial on the pretrain-finetune paradigm for natural language processing}.{\BBCQ}
\newblock
\APACjournalVolNumPages{arXiv:2403.02504}{}{}{}.
\PrintBackRefs{\CurrentBib}

\bibitem [\protect \citeauthoryear {%
Y.~Wang%
, Tian%
, Yazar%
, Ones%
\BCBL {}\ \BBA {} Landers%
}{%
Y.~Wang%
\ \protect \BOthers {.}}{%
{\protect \APACyear {2022}}%
}]{%
bag_of_words}
\APACinsertmetastar {%
bag_of_words}%
\begin{APACrefauthors}%
Wang, Y.%
, Tian, J.%
, Yazar, Y.%
, Ones, D\BPBI S.%
\BCBL {}\ \BBA {} Landers, R\BPBI N.%
\end{APACrefauthors}%
\unskip\
\newblock
\APACrefYearMonthDay{2022}{}{}.
\newblock
{\BBOQ}\APACrefatitle {Using Natural Language Processing and Machine Learning to Replace Human Content Coders} {Using natural language processing and machine learning to replace human content coders}.{\BBCQ}
\newblock
\APACjournalVolNumPages{Psychological Methods}{}{}{}.
\PrintBackRefs{\CurrentBib}

\bibitem [\protect \citeauthoryear {%
Y.~Wang%
, Yuan%
\BCBL {}\ \BBA {} Luo%
}{%
Y.~Wang%
\ \protect \BOthers {.}}{%
{\protect \APACyear {2015}}%
}]{%
tweets_china}
\APACinsertmetastar {%
tweets_china}%
\begin{APACrefauthors}%
Wang, Y.%
, Yuan, J.%
\BCBL {}\ \BBA {} Luo, J.%
\end{APACrefauthors}%
\unskip\
\newblock
\APACrefYearMonthDay{2015}{}{}.
\newblock
{\BBOQ}\APACrefatitle {America Tweets China: A Fine-Grained Analysis Of The State And Individual Characteristics Regarding Attitudes Towards China} {America tweets china: A fine-grained analysis of the state and individual characteristics regarding attitudes towards china}.{\BBCQ}
\newblock
\APACjournalVolNumPages{IEEE International Conference on Big Data}{}{}{}.
\PrintBackRefs{\CurrentBib}

\bibitem [\protect \citeauthoryear {%
Wei%
, Tay%
\BCBL {}\ \protect \BOthers {.}}{%
Wei%
, Tay%
\BCBL {}\ \protect \BOthers {.}}{%
{\protect \APACyear {2022}}%
}]{%
emerging_capabilities}
\APACinsertmetastar {%
emerging_capabilities}%
\begin{APACrefauthors}%
Wei, J.%
, Tay, Y.%
, Bommasani, R.%
, Raffel, C.%
, Zoph, B.%
, Borgeaud, S.%
\BDBL {}Fedus, W.%
\end{APACrefauthors}%
\unskip\
\newblock
\APACrefYearMonthDay{2022}{}{}.
\newblock
{\BBOQ}\APACrefatitle {Emergent Abilities of Large Language Models} {Emergent abilities of large language models}.{\BBCQ}
\newblock
\APACjournalVolNumPages{Transactions on Machine Learning Research}{}{}{}.
\PrintBackRefs{\CurrentBib}

\bibitem [\protect \citeauthoryear {%
Wei%
, Wang%
\BCBL {}\ \protect \BOthers {.}}{%
Wei%
, Wang%
\BCBL {}\ \protect \BOthers {.}}{%
{\protect \APACyear {2022}}%
}]{%
wei2022chain}
\APACinsertmetastar {%
wei2022chain}%
\begin{APACrefauthors}%
Wei, J.%
, Wang, X.%
, Schuurmans, D.%
, Bosma, M.%
, Ichter, B.%
, Xia, F.%
\BDBL {}Zhou, D.%
\end{APACrefauthors}%
\unskip\
\newblock
\APACrefYearMonthDay{2022}{}{}.
\newblock
{\BBOQ}\APACrefatitle {Chain-of-Thought Prompting Elicits Reasoning in Large Language Models} {Chain-of-thought prompting elicits reasoning in large language models}.{\BBCQ}
\newblock
\BIn{} \APACrefbtitle {Advances in Neural Information Processing Systems 35 (NeurIPS 2022) Main Conference Track.} {Advances in neural information processing systems 35 (neurips 2022) main conference track.}
\PrintBackRefs{\CurrentBib}

\bibitem [\protect \citeauthoryear {%
Widmann%
\ \BBA {} Wich%
}{%
Widmann%
\ \BBA {} Wich%
}{%
{\protect \APACyear {2023}}%
}]{%
german_compare_emotion}
\APACinsertmetastar {%
german_compare_emotion}%
\begin{APACrefauthors}%
Widmann, T.%
\BCBT {}\ \BBA {} Wich, M.%
\end{APACrefauthors}%
\unskip\
\newblock
\APACrefYearMonthDay{2023}{}{}.
\newblock
{\BBOQ}\APACrefatitle {Creating and Comparing Dictionary, Word Embedding, and Transformer-Based Models to Measure Discrete Emotions in German Political Text} {Creating and comparing dictionary, word embedding, and transformer-based models to measure discrete emotions in german political text}.{\BBCQ}
\newblock
\APACjournalVolNumPages{Political Analysis}{}{}{}.
\PrintBackRefs{\CurrentBib}

\bibitem [\protect \citeauthoryear {%
Zhang%
\ \protect \BOthers {.}}{%
Zhang%
\ \protect \BOthers {.}}{%
{\protect \APACyear {2021}}%
}]{%
monitor_depression}
\APACinsertmetastar {%
monitor_depression}%
\begin{APACrefauthors}%
Zhang, Y.%
, Lyu, H.%
, Liu, Y.%
, Zhang, X.%
, Wang, Y.%
\BCBL {}\ \BBA {} Luo, J.%
\end{APACrefauthors}%
\unskip\
\newblock
\APACrefYearMonthDay{2021}{}{}.
\newblock
{\BBOQ}\APACrefatitle {Monitoring Depression Trends on Twitter During the COVID-19 Pandemic: Observational Study} {Monitoring depression trends on twitter during the covid-19 pandemic: Observational study}.{\BBCQ}
\newblock
\APACjournalVolNumPages{JMIR Infodemiology}{}{}{}.
\PrintBackRefs{\CurrentBib}

\bibitem [\protect \citeauthoryear {%
Zhong%
, Ding%
, Liu%
, Du%
\BCBL {}\ \BBA {} Tao%
}{%
Zhong%
\ \protect \BOthers {.}}{%
{\protect \APACyear {2023}}%
}]{%
chatgpt_vs_bert}
\APACinsertmetastar {%
chatgpt_vs_bert}%
\begin{APACrefauthors}%
Zhong, Q.%
, Ding, L.%
, Liu, J.%
, Du, B.%
\BCBL {}\ \BBA {} Tao, D.%
\end{APACrefauthors}%
\unskip\
\newblock
\APACrefYearMonthDay{2023}{}{}.
\newblock
{\BBOQ}\APACrefatitle {Can ChatGPT Understand Too? A Comparative Study on ChatGPT and Fine-tuned BERT} {Can chatgpt understand too? a comparative study on chatgpt and fine-tuned bert}.{\BBCQ}
\newblock
\APACjournalVolNumPages{arXiv:2302.10198}{}{}{}.
\PrintBackRefs{\CurrentBib}

\bibitem [\protect \citeauthoryear {%
Ziems%
\ \protect \BOthers {.}}{%
Ziems%
\ \protect \BOthers {.}}{%
{\protect \APACyear {2024}}%
}]{%
llm_transform_css}
\APACinsertmetastar {%
llm_transform_css}%
\begin{APACrefauthors}%
Ziems, C.%
, Held, W.%
, Shaikh, O.%
, Chen, J.%
, Zhang, Z.%
\BCBL {}\ \BBA {} Yang, D.%
\end{APACrefauthors}%
\unskip\
\newblock
\APACrefYearMonthDay{2024}{{\APACmonth{03}}}{}.
\newblock
{\BBOQ}\APACrefatitle {Can Large Language Models Transform Computational Social Science?} {Can large language models transform computational social science?}{\BBCQ}
\newblock
\APACjournalVolNumPages{Computational Linguistics}{50}{1}{}.
\PrintBackRefs{\CurrentBib}

\end{thebibliography}
\end{document}